\pdfoutput=1

\documentclass[11pt]{article}

\usepackage{acl}

\usepackage{times}
\usepackage{latexsym}
\usepackage{graphicx}
\usepackage{multirow}
\usepackage{amsmath}
\usepackage{diagbox}
\usepackage{todonotes}
\usepackage{multicol}
\usepackage{adjustbox}
\usepackage{tabularx}
\usepackage{booktabs}
\usepackage{paralist}
\usepackage{float}
\usepackage{hyperref}

\newcommand{\marktext}[2]{\adjustbox{bgcolor=#1}{\strut #2}}

\usepackage[T1]{fontenc}

\usepackage[utf8]{inputenc}

\usepackage{microtype}

\usepackage{inconsolata}

\usepackage[most]{tcolorbox}
\usepackage{xcolor}
\usepackage{colortbl}
\definecolor{myorange}{HTML}{fa8072}
\definecolor{mygreen}{HTML}{9DC183}

\let\svthefootnote\thefootnote
\newcommand\freefootnote[1]{%
  \let\thefootnote\relax%
  \footnotetext{#1}%
  \let\thefootnote\svthefootnote%
}

\title{Fact-Checking the Output of Large Language Models\\ via Token-Level Uncertainty Quantification}

\author{
\bf Ekaterina Fadeeva\textsuperscript{3,4 $\diamondsuit$}\quad
Aleksandr Rubashevskii\textsuperscript{1,3 $\diamondsuit$}\quad
Artem Shelmanov\textsuperscript{1 $\diamondsuit$}\\
\bf Sergey Petrakov\textsuperscript{3}\enspace 
Haonan Li\textsuperscript{1}\enspace 
Hamdy Mubarak\textsuperscript{7}\enspace 
Evgenii Tsymbalov\textsuperscript{8}\enspace
Gleb Kuzmin\textsuperscript{2,5} \\ 
\bf Alexander Panchenko\textsuperscript{2,3}\quad
Timothy Baldwin\textsuperscript{1,6}\quad
Preslav Nakov\textsuperscript{1}\quad
Maxim Panov\textsuperscript{1}\\
\textsuperscript{1}MBZUAI\quad 
\textsuperscript{2}AIRI\quad
\textsuperscript{3}Center for Artificial Intelligence Technology \quad
\textsuperscript{4}HSE University\quad \\
\textsuperscript{5}FRC CSC RAS\quad
\textsuperscript{6}The University of Melbourne\enspace 
\textsuperscript{7}QCRI\enspace
\textsuperscript{8}Independent Researcher
\\
\href{mailto:ekaterina.fadeeva@skol.tech}{\{ekaterina.fadeeva, sergey.petrakov\}@skol.tech} ~~
\href{mailto:kuzmin@airi.net}{\{kuzmin, panchenko\}@airi.net} \\
\href{mailto:artem.shelmanov@mbzuai.ac.ae}{\{aleksandr.rubashevskii, artem.shelmanov, haonan.li\}@mbzuai.ac.ae} \\
\href{mailto:maxim.panov@mbzuai.ac.ae}{\{maxim.panov, timothy.baldwin, preslav.nakov\}@mbzuai.ac.ae} \quad
\href{mailto:hmubarak@hbku.edu.qa}{hmubarak@hbku.edu.qa}
}

\newcommand\blankfootnote[1]{%
  \let\thefootnote\relax\footnotetext{#1}%
  \let\thefootnote\svthefootnote%
}

\usepackage{dsfont}

\usepackage{multirow}
\usepackage{xspace}
\usepackage{cleveref}
\newcommand{\multirowcell}[1]{\begin{tabular}[c]{@{}l@{}}#1\end{tabular}}
\newcommand{\ex}[1]{\textit{#1}\xspace}

\begin{document}
\maketitle

\begin{abstract}
  Large language models (LLMs) are notorious for hallucinating, i.e., producing erroneous claims in their output. Such hallucinations can be dangerous, as  occasional factual inaccuracies in the generated text might be obscured by the rest of the output being generally factually correct, making it extremely hard for the users to spot them. Current services that leverage LLMs usually do not provide any means for detecting unreliable generations. Here, we aim to bridge this gap. In particular, we propose a novel fact-checking and hallucination detection pipeline based on token-level uncertainty quantification. Uncertainty scores leverage information encapsulated in the output of a neural network or its layers to detect unreliable predictions, and we show that they can be used to fact-check the atomic claims in the LLM output. Moreover, we present a novel token-level uncertainty quantification method that removes the impact of uncertainty about what claim to generate on the current step and what surface form to use. Our method Claim Conditioned Probability (CCP) measures only the uncertainty of a particular claim value expressed by the model. Experiments on the task of biography generation demonstrate strong improvements for CCP compared to the baselines for seven LLMs and four languages. Human evaluation reveals that the fact-checking pipeline based on uncertainty quantification is competitive with a fact-checking tool that leverages external knowledge.
\end{abstract}



\section{Introduction}
\label{sec:introduction}

\freefootnote{$\diamondsuit$ Equal contribution}


Large language models (LLMs) have become a ubiquitous and versatile tool for addressing a variety of natural language processing (NLP) tasks. 
People use these models for tasks including information search~\cite{Sun2023IsCG}, to ask medical questions~\cite{thirunavukarasu2023large}, or to generate new content~\cite{sun-etal-2023-evaluating}. Recently, there has been a notable shift in user behavior, indicating an increasing reliance on and trust in LLMs as primary information sources, often surpassing traditional channels.
However, 
a significant challenge with the spread of these models is their tendency to produce ``hallucinations'', i.e.,~factually incorrect generations that contain misleading information~\cite{bang-etal-2023-multitask,dale-etal-2023-detecting}. This is a side-effect of the way modern LLMs are designed and trained~\cite{kalai2023calibrated}.

LLM hallucinations are a major concern because the deceptive content at the surface level can be highly coherent and persuasive. 
Common examples include the creation of fictitious biographies or the assertion of unfounded claims. The danger is that a few occasional false claims might be easily obscured by a large number of factual statements, making it extremely hard for people to spot them. 
As hallucinations in LLM outputs are hard to eliminate completely, users of such systems could be informed via highlighting some potential caveats in the text, and this is where our approach can help.

Fact-checking is a research direction that addresses this problem. 
It is usually approached using complex systems that leverage external knowledge sources~\cite{guo-etal-2022-survey,nakov2021clef,wadden-etal-2020-fact}. This introduces problems related to the incomplete nature of such sources and notable overhead in terms of storing the knowledge. We argue that information about whether a generation is a hallucination is encapsulated in the model output itself, and can be extracted using uncertainty quantification (UQ)~\cite{gal2016uncertainty,kotelevskii2022nonparametric,vazhentsev-etal-2022-uncertainty,vazhentsev-etal-2023-hybrid}. This avoids implementing complex and expensive fact-checking systems that require additional computational overhead and rely on external resources.

Prior work has mainly focused on quantification of uncertainty for the whole generated text and been mostly limited to tasks such as machine translation~\cite{malinin2020uncertainty}, question answering~\cite{kuhn2023semantic}, and text summarization~\cite{van-der-poel-etal-2022-mutual}. However, the need for an uncertainty score for only a part of the generation substantially complicates the problem. We approach it by leveraging token-level uncertainty scores and aggregating them into claim-level scores. Moreover, we introduce a new token-level uncertainty score, namely claim-conditioned probability (CCP), which demonstrates confident improvements over several baselines for seven LLMs and four languages. 

To the best of our knowledge, there is no previous work that has investigated the quality of claim-level UQ techniques for LLM generation. Therefore, for this purpose, we construct a novel benchmark based on fact-checking of biographies of individuals generated using a range of LLMs.  Note that different LLMs produce different outputs, which generally have higher variability than, e.g., outputs in such tasks as machine translation or question answering. Therefore, we compare the predictions and uncertainty scores to the results of an automatic external fact-checking system FactScore~\cite{min2023factscore}. Human evaluation verifies that our constructed benchmark based on FactScore can adequately 
evaluate the performance of the uncertainty scores.



Our contributions are as follows:
\begin{compactitem}
  \item We propose a \textit{novel framework} for fact-checking LLM generations using token-level uncertainty quantification.
  We provide a procedure for efficiently estimating the uncertainty of atomic claims generated by a white-box model and highlighting potentially deceptive fragments by mapping them back to the original response.

  \item We propose a \textit{novel method} for token-level uncertainty quantification that outperforms baselines and can be used as a plug-in in a fact-checking framework.

  \item We design a \textit{novel approach to evaluation} of token-level UQ methods for white-box LLMs based on fact-checking, which can be applied to other white-box LLMs.

  \item We provide an \textit{empirical and ablation analysis} of the method for fact-checking of LLM generations, and find that the uncertainty scores we produce can help to spot claims with factual errors for seven LLMs over four languages: English, Chinese, Arabic, and Russian.

  \item The method is implemented as a part of the LM-Polygraph library \cite{fadeeva-etal-2023-lm}. All the code and data for experiments is publicly available\footnote{\url{https://github.com/IINemo/lm-polygraph}}.
\end{compactitem}



\setlength{\columnsep}{0.3pc}

\begin{figure*}[h]
	\centering
	\scriptsize
 \begin{multicols}{2}

Maximum Probability
 
	\begin{tcolorbox}[size=fbox,
		title=\,{\includegraphics[width=1.5em]{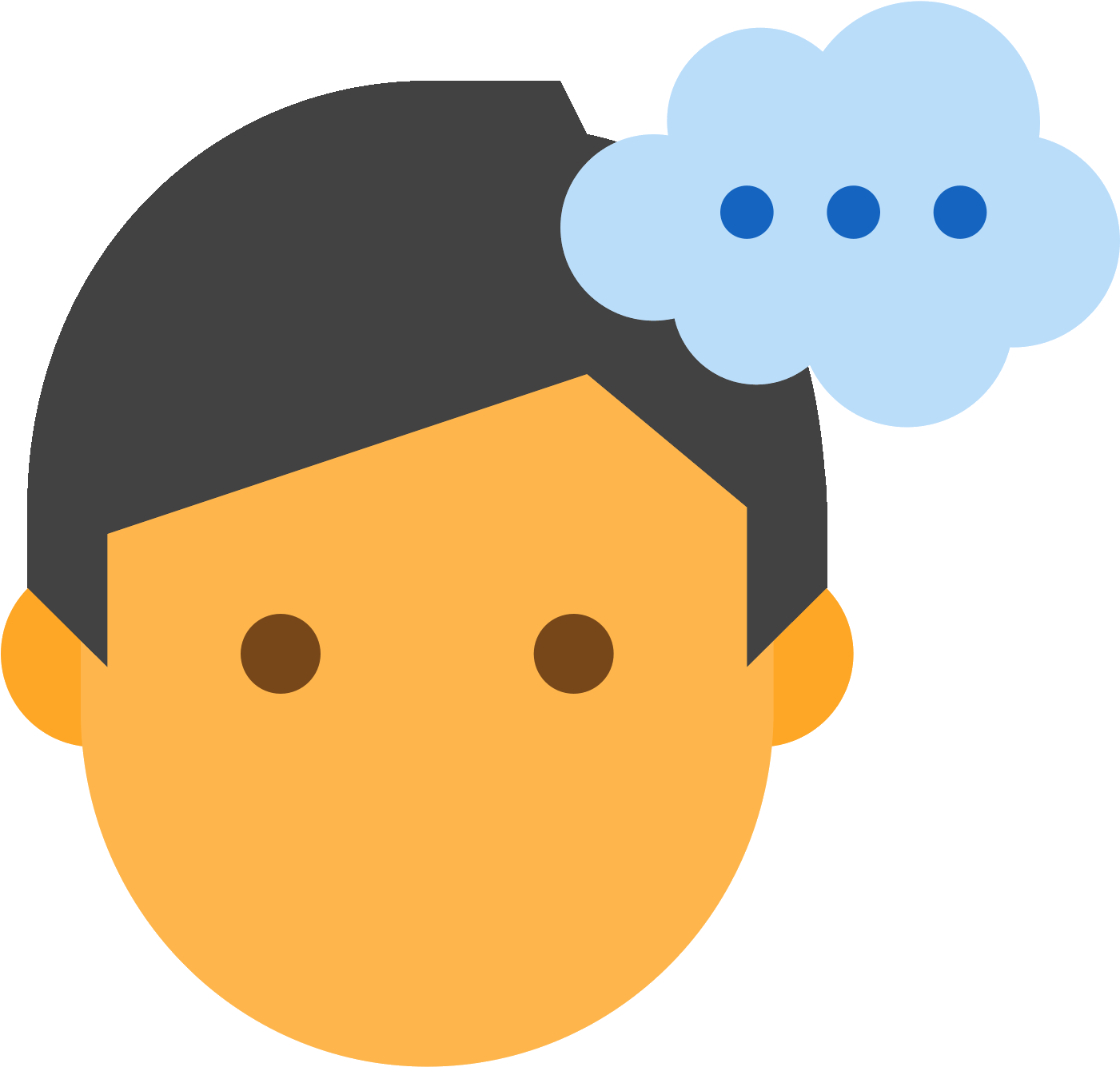}\,\,\,Tell me a bio of Madonna.},
		colback=gray!2!white,
		colframe=cyan!15!white,
            coltitle=black,
		width=32em,
		sharp corners=northwest,
		]
			\includegraphics[width=1.5em]{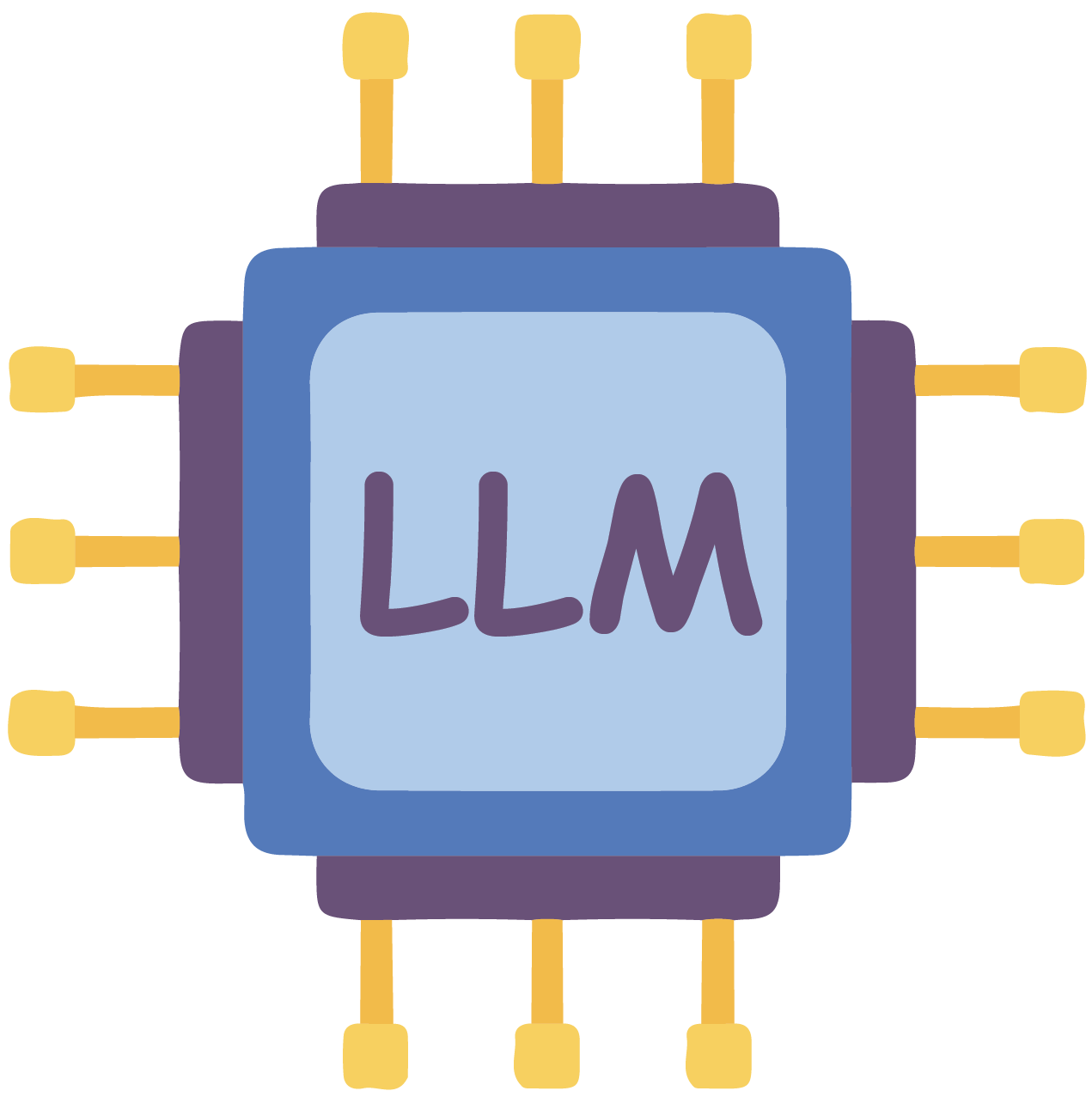}
			 \,\,\,Madonna is an American singer, songwriter, actress, and businesswoman. She was born on August 16, 1958, in Bay City, Michigan, and is \marktext{myorange}{best known for her provocative image and music}. Madonna has sold over 300 million records worldwide, making her one of the best-selling music artists of all time. She has won numerous awards, including 20 Grammy Awards, a Golden Globe Award, and an Emmy Award. Madonna is also known for her philanthropic work and her activism on various social issues.

	\end{tcolorbox}

\columnbreak

Claim-Conditioned Probability (ours)

\begin{tcolorbox}[size=fbox,
		title=\,{\includegraphics[width=1.5em]{tex/img/user-icon.png}\,\,\,Tell me a bio of Madonna.},
		colback=gray!2!white,
		colframe=cyan!15!white,
            coltitle=black,
		width=32em,
		sharp corners=northwest,
		]
			\includegraphics[width=1.5em]{tex/img/llm-icon.png}
			\,\,\,Madonna is an American singer, songwriter, actress, and businesswoman. She was born on August 16, 1958, in Bay City, Michigan, and is best known for her provocative image and music. Madonna has sold over 300 million records worldwide, making her one of the best-selling music artists of all time. She has won numerous awards, including \marktext{myorange}{20 Grammy Awards}, a Golden Globe Award, and an Emmy Award. Madonna is also known for her philanthropic work and her activism on various social issues.
	\end{tcolorbox}
 
 \end{multicols}
	    \caption{Visual comparison of our Claim-Conditioned Probability method to the Maximum Probability baseline. CCP accurately identifies the incorrectly specified number of awards (in \colorbox{myorange}{red}),  whereas Maximum Probability erroneously highlights the claim that is actually correct.}
	\label{fig:token-level_comparison_app3}
\end{figure*}

\section{Related Work}
\label{sec:related_work}

\subsection{Fact-Checking LLM Generations and Detecting Hallucinations}

The problem of hallucinations has made fact-checking of LLM outputs a prominent topic in the research community and resulted in a surge of publications on the topic.
\citet{chern2023factool} present Factool -- a task and domain agnostic framework for hallucination detection that leverages GPT for claim extraction and verification.
\citet{manakul2023selfcheckgpt} suggest to sample multiple outputs from black-box LLMs and evaluate how similar the sampled responses are using the external model. 
\citet{varshney2023stitch} detect LLM hallucinations by extracting key parts of output using an external model and  estimating their uncertainty based on logits. The most uncertain parts are verified using an external knowledge source.
\citet{pan-etal-2023-fact} propose to fact-check complex statements by decomposing them into simpler subtasks and generating reasoning programs to verify these statements. 
\citet{min2023factscore} present a methodology for evaluating long LLM-generated texts by decomposing them into simple atomic statements and further verifying them against some knowledge source.
Several subsequent works further optimize components of the knowledge-based fact-checking pipelines~\citep{loki,wang2024factcheckbench}.

In contrast to previous work that leverages external knowledge sources for fact-checking (a database or another LLM), our work is the first to investigate token-level UQ methods for this task using the LLM and its outputs only.




\subsection{Uncertainty Quantification of LLM Generations}

UQ techniques for LLM generation can be classified into five major categories~\cite{fadeeva-etal-2023-lm}.
Information-based methods leverage the probability distribution of generated tokens and usually do not require any additional models. In this category, we can include methods such as perplexity~\cite{fomicheva-etal-2020-unsupervised}, mean token entropy~\cite{fomicheva-etal-2020-unsupervised},
point-wise mutual information (PMI)~\cite{takayama-arase-2019-relevant}, and conditional PMI~\cite{van-der-poel-etal-2022-mutual}. 

Another category of methods is based on density estimation of latent instance representations. A typical example in this category is Mahalanobis distance~\cite{lee_simple2018} and its various modifications~\cite{ren2023outofdistribution,vazhentsev-etal-2023-efficient}. 
The disadvantage of such methods is the need for access to the LLM training data in order to fit external density models, which is problematic for most general-purpose LLMs.

Ensembling and Monte Carlo dropout methods are based on the lexical diversity of multiple outputs sampled from one or multiple versions of LLMs for a single query~\cite{malinin2020uncertainty,fomicheva-etal-2020-unsupervised}. Their main drawback is that they require many predictions, which makes them too computationally and memory intensive for practical purposes. Additionally, it is difficult to apply these techniques to quantify uncertainty of text fragments such as claims, as different samples may seriously diverge.


It has recently been shown that LLMs can reflexively estimate the confidence of their generations simply by asking themselves about the truthfulness of their output~\citep{kadavath2022language}. This can work better than analyzing the probability distribution for the original prediction, but requires a second pass of inference, feeding the original output as a part of the query.


Finally, there is a group of methods that leverages the diversity of meanings that the LLM generates for a given query. This group includes semantic entropy~\cite{kuhn2023semantic} and various scores based on the analysis of the similarity matrix between outputs~\cite{lin2023generating}. 

UQ methods can also be classified into white-box and black-box \cite{lin2023generating}  approaches, depending on the required access to LLM itself and its outputs. Black-box techniques do not require any other input except generated texts. 





The method in our work can be attributed to the information-based group, and can be applied only to white-box LLMs because it requires access to the probability distribution of the generated tokens. Compared to other techniques, it offers a novel approach to post-processing the probability distribution and is specifically designed to quantify uncertainty of output fragments, such as atomic claims and individual words.

\section{Fact-Checking Pipeline}
  The fact-checking pipeline (see Figure~\ref{fig:biography_scheme_general}) starts with splitting a generated text into atomic claims, e.g. using a much smaller model fine-tuned for this particular task. For experimental evaluations in this work, we follow the FactScore approach~\cite{min2023factscore}, where splitting is implemented via the OpenAI Chat API.


  Each atomic claim is matched against the sequence of tokens in the original text with the corresponding probability distributions. Then we calculate token-level uncertainty scores and aggregate them into the claim-level uncertainty. 

  Finally, the claim-level uncertainty scores are compared against a threshold obtained on a validation set to determine whether the claim should be highlighted for the end-user as unreliable. Individual tokens can be a part of multiple atomic claims. If the token belongs to a reliable and unreliable claim at the same time, it is not highlighted. An example visualization is presented in Figure~\ref{fig:token-level_comparison_app3}. 
  

\begin{figure*}[t]
    \centering
    \includegraphics[scale=0.61,trim={0 2pt 0 0},clip]{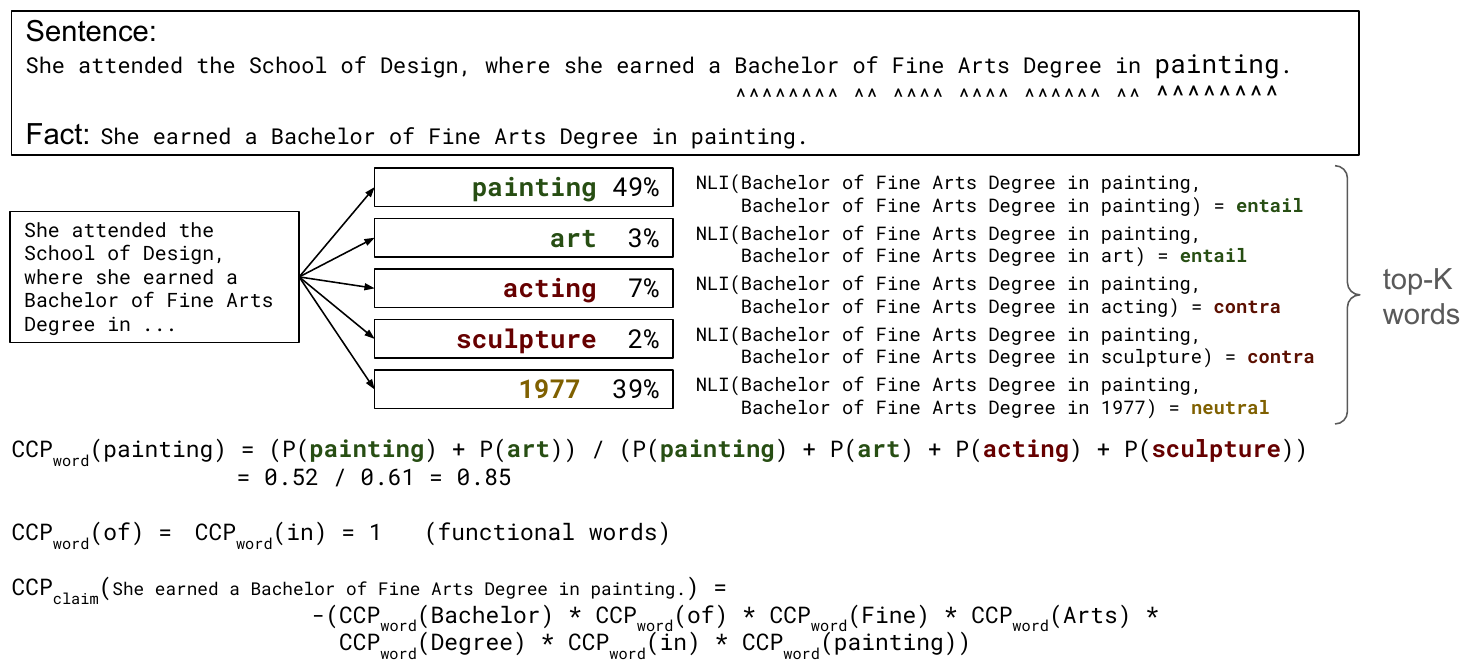}
    \caption{Example of CCP calculation for the word \ex{painting} in a Vicuna 13b generation. 
    }
  \label{fig:nli_ex1}
  \end{figure*}

\section{Uncertainty Quantification}

  In this section, we first provide background on common UQ methods that can be used at the token level, then delve into our \textbf{Claim-Conditioned Probability (CCP)} token-level method, and finally describe how token-level uncertainties are aggregated into a claim-level score.

  Autoregressive language models generate text token by token. In this work, we will operate on the level of words and without loss of generality suppose that the autoregressive distribution at each step generates the random word $X_j \sim P(\cdot \mid x_{<j})$, where $x_{<j}$ is the text generated before the word at position $j$. We also denote by $x_j$ the generated word at position $j$ and by $x_{1:j} = x_{<j} \circ x_j$ a text composed of words at positions $1$ to $j$. For example, in the case of greedy generation, $x_j = \arg\max_x{P(x \mid x_{<j})}$, where $x_j$ is the most probable realization of $X_j$. Let us also denote by $C$ a set of indices of words corresponding to a particular atomic claim. 

\subsection{Claim-Level UQ Baselines}
\label{sec:baselines}
  We note that for UQ to be practical, it needs to be fast. Therefore, we do not consider methods such as deep ensembles~\cite{DeepEnsemble}, due to their significant computational overhead.

\textbf{Maximum Probability} represents a basic approach to UQ, where we simply treat the probability of the most likely generation as a confidence score:
  \begin{equation}
    MP(C) = 1 - \prod\nolimits_{j \in C} P (x_{j} \mid x_{<j}).
  \end{equation}
  %

\textbf{Perplexity} is a common metric used to evaluate the performance of LLMs. Lower perplexity indicates that a model's probability distribution better predicts a sample. It is computed
 as the average negative log probability of generated tokens that belong to the claim $C$:
  \begin{equation}
  \small
    Perp(C) = \exp\left(-\frac{1}{|C|} \sum\nolimits_{j \in C} \log P (x_{j} \mid x_{<j})\right).
  \end{equation}

\textbf{Maximum Entropy} of a token in the claim: 
  \begin{equation}
    Ent(C) = \max\nolimits_{j \in C} \mathcal{H}(\cdot \mid x_{<j}),
  \end{equation}
  where $\mathcal{H}(\cdot \mid x_{<j})$ is the entropy of the autoregressive distribution of the current token. Preliminary experiments indicated that simply getting the maximum of the token entropies in a claim noticeably outperforms other aggregation techniques like average or minimum. It is also generally a slightly better baseline than perplexity.

\textbf{P(True),} similar to~\cite{kadavath2022language}, measures the uncertainty of the claim by asking the LLM itself whether the generated claim is true or not. The confidence is the probability of the first generated token $y_1$ being equal to ``True'':
\begin{equation}
  P(\text{True}) = 1 - P(y_1 = \text{``True''}).
\end{equation}
While some work has reported that this technique outperforms other baselines, the big drawback is that one needs to run the original LLM twice.

\subsection{Claim-Conditioned Probability}
  In this subsection, we propose a novel method for token- and claim-level uncertainty quantification.

\subsubsection{Motivation and Theoretical Background}
  When an LLM generates an output, it faces various types of uncertainty reflected in the token distribution of the current generation step (see Figure~\ref{fig:nli_ex2} for an example). We identify three distinct types of uncertainty:
  
    \textbf{(1) Claim type/order uncertainty}: What claim to generate on the current step? For example, on the current step, an LLM might hesitate between generating a year of graduation of a person and a field of study. A different order of claims, missing claims, or different types of generated claims do not make produced text less factual. Therefore, when performing fact-checking, we should not take this type of uncertainty into account.

    \textbf{(2) Surface form uncertainty}: What synonyms or hypernyms to use when generating a claim (e.g. ``art'' or ``painting'')? Different surface forms also do not make the text less factual --- they might change the style, but not the underlying meaning of the text. Therefore, this type of uncertainty is also not relevant for fact-checking.
    

    \textbf{(3) Claim uncertainty}: What specific piece of information to relay for a particular claim type?  For example, an LLM might not be sure which field of study to generate, producing a token distribution with multiple highly-probable variants, such as ``painting'', ``acting'', ``sculpture''.
    Similarly, for the year of graduation, an LLM might produce a distribution with various potential years. This uncertainty is relevant for fact-checking, because if the model is not sure about the information it relays, there might be a high chance of a factual mistake.

  Two out of the three types of uncertainty are irrelevant for fact-checking and only introduce noise into the final score. We propose a new UQ method that ignores the first two types of uncertainty and focuses only on the third one, namely \textbf{Claim-Conditioned Probability (CCP)}:
  \begin{equation}
  \small
    \!\!\!\! CCP(x_j) = P\bigl(\text{Meaning}(x_{1:j}) | x_{<j}, \text{ClaimType}(x_{1:j})\bigr). \!\!\!
  \label{eqn:ccp_general}
  \end{equation}
  Here, $\text{ClaimType}(x_{1:j})$ represents a claim type of the generated sequence $x_{1:j}$ and $\text{Meaning}(x_{1:j})$ is a function that maps $x_{j}$ into its meaning given the previous words in a sentence $x_{<j}$,
  so that various surface forms with a similar meaning for $x_j$ are mapped to a single categorical variable. 

  Conditional probability can be rewritten using unconditional probabilities:
  \begin{align*}
    & P\bigl(\text{Meaning}(x_{1:j}) \mid \text{ClaimType}(x_{1:j}), x_{<j}\bigr) \\
    & = \frac{P(\text{Meaning}(x_{1:j}), \text{ClaimType}(x_{1:j}) \mid x_{<j})}{P(\text{ClaimType}(x_{1:j}) \mid x_{<j})}.
  \label{eqn:ccp_genral_2}
  \end{align*}
  Assuming that each meaning in a word distribution can correspond to only a single claim type, the joint probability is the same as the meaning probability $P(\text{Meaning}(x_{1:j}),  \text{ClaimType}(x_{1:j}) \mid x_{<j}) = P(\text{Meaning}(x_{1:j}) \mid x_{<j})$.

  Meaning probability in turn sums from the probabilities of word alternatives $x^k_j$ that correspond to the same meaning: $P(\text{Meaning}(x_{1:j}) \mid x_{<j}) = \sum_{x^k_j \in M(x_j)}{}{P(x^k_j \mid x_{<j})}$, where we say that $x^k_j \in M(x_j)$ if $\text{Meaning}(x_{1:j}) = \text{Meaning}(x_{<j} \circ x^k_j)$.

  In the same way, the probability of a claim type sums from probabilities of all meanings and transitionally from probabilities of words that correspond to the particular claim type: $P\bigl(\text{ClaimType}(x_{1:j})\bigr) = \sum_{x^l_j \in CT(x_j)}{}{P(x^l_j \mid x_{<j})}$, where we denote by $x^l_j \in CT(x_j)$ an event such that $\text{ClaimType}(x_{1:j}) = \text{ClaimType}(x_{<j} \circ x^l_j)$. Therefore, equation~\eqref{eqn:ccp_general} can be rewritten as follows:
  \begin{equation}
    CCP(x_j) = \frac{\sum_{x^k_j \in M(x_{j})}{}{P(x^k_j \mid x_{<j})}}{\sum_{x^l_j \in CT(x_{j})}{}{P(x^l_j \mid x_{<j})}}.
  \label{eqn:ccp_general_3}
  \end{equation}
  The meaning function and the construction of the set of words that belong to the same claim type can be implemented in various ways. We outline our approach in \Cref{sec:implementation}.

  Previously-proposed UQ methods have partially accounted for some of the types of uncertainty described above. For example, semantic entropy~\cite{kuhn2023semantic} accounts for uncertainty in semantically-equivalent groups, which helps to alleviate the surface-form uncertainty. However, in our method, we additionally remove the impact of claim-type uncertainty.

\subsubsection{Implementation}
\label{sec:implementation}







  We implement CCP using NLI at the word level. We compare the original claim and the claim, in which the target word is replaced by its alternatives from the autoregressive distribution. 
    \vspace{1mm}
  
  The distribution $X_j$ at the position $j$ is approximated by top-$K$ alternatives $\{x^k_j\}_{k=1}^K$  
  with $x^1_j \equiv x_j$. 
  We replace $x_j$ with its alternatives $x_j^k$ and obtain new instances $x_{<j} \circ x^k_j, k = 1, \dots, K$. 
  Each new instance is compared against the original prediction $x_{1:j} = x_{<j} \circ x_j$ using an NLI model. 
  We define $\texttt{NLI}(x_j^k, x_j) := \texttt{NLI}(x_{<j} \circ x^k_j, x_{1:j})$, where $\texttt{NLI}(x_{<j} \circ x^k_j, x_{1:j})$ means application of the NLI model to the text fragments $x_{<j} \circ x^k_j$ and $x_{1:j}$. 

The outcome of the NLI procedure is one of three labels: entail (`e'), contradict (`c'), or neutral (`n'). If the new instance entails the original prediction $\texttt{NLI}(x_j^k, x_j) = \textrm{`e'}$, then we consider $x_{<j} \circ x^k_j$ has the same meaning with $x_{1:j}$ ($x^k_j \in M(x_{j})$) and corresponds to the same claim type ($x^k_j \in CT(x_{j})$).

If the new instance contradicts the original prediction $\texttt{NLI}(x_j^k, x_j) = \textrm{`c'}$, then we consider $x_{<j} \circ x^k_j$ has a different meaning with $x_{1:j}$ ($x^k_j \notin M(x_{j})$), but corresponds to the same claim type ($x^k_j \in CT(x_{j})$). Otherwise, if the new instance is neutral w.r.t. the original prediction $\texttt{NLI}(x_j^k, x_j) = \textrm{`n'}$, then we consider that $x_{<j} \circ x^k_j$ does not correspond to the same claim type as $x_{1:j}$ ($x^k_j \notin CT(x_{j})$). Thus, equation~\eqref{eqn:ccp_general_3} for CCP can be written as follows:
  \begin{small}
  \begin{equation*}
    CCP_{\textit{word}}(x_j) =
      \dfrac{\sum_{k: \texttt{NLI}(x^k_j, x_j) = \texttt{`e'}} P(x^k_j \mid x_{<j})} {\sum_{k: \texttt{NLI}(x^k_j, x_j) \in \left\{ \texttt{`e'}, \texttt{`c'}\right\}} P(x^k_j \mid x_{<j} )}.
  \label{eqn:ccp}
  \end{equation*}
  \end{small} 
For practical considerations, we consider that CCP for function words is always equal to 1. In our experiments, we base this determination on the stop word list from NLTK~\cite{nltk}. 

  We note that most transformer LLMs generate sub-word tokens instead of whole words. To obtain distributions for whole words, we generate one or multiple tokens using beam search with $K$ beams.



  To obtain CCP-based claim-level uncertainty, we simply take the product of CCPs of each word from the claim $C$: 
  \begin{equation}
    CCP_{\textit{claim}}(C) = 1 - \prod\nolimits_{j \in C} CCP_{\textit{word}}(x_j).
  \end{equation}
An example of calculating the CCP for a claim is presented in Figure~\ref{fig:nli_ex1}. Other detailed examples of CCP calculation are available in Appendix~\ref{sec:ccp_explained}.


\section{Benchmark for Evaluation of Claim-Level UQ Methods}
\label{sec:dataset}

  We evaluate claim-level UQ techniques and their ability to spot hallucinations on the task of generating biographies. 
   In relevant previous work \cite{manakul2023selfcheckgpt}, the authors generate biographies with one LLM (GPT-3), manually annotate sentences for factuality, and quantify uncertainty of a different ``proxy'' model. Factuality labels are then used to evaluate the quality of uncertainty scores. We argue that such an approach based on a proxy model introduces a big discrepancy between the generated text and what a proxy LLM actually wants to generate, which results in biased UQ evaluation results.
   To make the evaluation as close as possible to the real-world scenario, we allow unrestricted generation of biographies from LLMs.
   
  Unrestricted generation complicates automatic evaluation of the fact-checking pipeline, because obtaining gold standard annotation requires manual annotation of all outputs from each model.
  Therefore, in addition to manual annotation, we annotate claims in generated texts automatically using FactScore -- a fact-checking tool~\cite{min2023factscore}, which has access to an external knowledge source. 
  Using FactScore, enables completely automatic evaluation and allows us to scale up experiments.

  We generate LLM responses in English, Chinese, Arabic, and Russian to $100$ biography prompts.
  The typical biography prompt is \ex{Give me a biography for <person name>} in different languages. The set of people was generated by asking GPT-4 to list the most famous people since $1900$. The maximum generation length is set to $256$ tokens. If the last sentence of the generation is unfinished (i.e.\ does not end with any punctuation), it is discarded.
  We generate responses for the following LLMs: (for English) Vicuna 13b~\cite{zheng2023judging}, Mistral 7b~\cite{jiang2023mistral}, Jais 13b~\cite{sengupta2023jais}, and GPT-3.5-turbo~\cite{ouyang2022training}; (for Chinese) Yi 6b~\cite{Yi2023}; (for Arabic) Jais 13b and GPT-4; (for Russian) Vikhr-instruct-0.2 7b \cite{nikolich2024vikhr}. 



  \begin{table*}[t]
\centering

\begin{tabular}{l|c|c|c|c}
\hline
\textbf{Model} & \textbf{Mistral 7b} & \textbf{Vicuna 13b} & \textbf{Jais 13b} & \textbf{GPT-3.5-turbo}\\
\hline
CCP (ours) & \textbf{0.66} \tiny{$\pm$ 0.03} & \textbf{0.66} \tiny{$\pm$ 0.04} & \textbf{0.71} \tiny{$\pm$ 0.05} & \textbf{0.58} \tiny{$\pm$ 0.04} \\
Maximum Prob. & 0.59 \tiny{$\pm$ 0.03} & 0.60 \tiny{$\pm$ 0.05} & 0.64 \tiny{$\pm$ 0.05} & 0.54 \tiny{$\pm$ 0.05} \\
Perplexity & 0.58 \tiny{$\pm$ 0.07} & 0.58 \tiny{$\pm$ 0.07} & 0.61 \tiny{$\pm$ 0.08} & 0.53 \tiny{$\pm$ 0.06} \\
Token Entropy & 0.60 \tiny{$\pm$ 0.02} & 0.60 \tiny{$\pm$ 0.06} & 0.63 \tiny{$\pm$ 0.06} & 0.53 \tiny{$\pm$ 0.03} \\
P(True) & 0.53 \tiny{$\pm$ 0.02} & 0.61 \tiny{$\pm$ 0.03} & 0.55 \tiny{$\pm$ 0.05} & 0.53 \tiny{$\pm$ 0.04} \\
\hline
\end{tabular}

\caption{\label{tab:roc-auc} ROC-AUC of claim-level UQ methods with FactScore labels as the ground truth (English). }
\end{table*}

\begin{table*}[t]
\centering

\begin{tabular}{l|c|c|c|c}
\hline
\textbf{Model} & \textbf{Mistral 7b} & \textbf{Vicuna 13b} & \textbf{Jais 13b} & \textbf{GPT-3.5-turbo}\\
\hline
CCP (ours) & \textbf{0.34} \tiny{$\pm$ 0.05} & \textbf{0.24} \tiny{$\pm$ 0.04} & \textbf{0.33} \tiny{$\pm$ 0.07} & 0.14 \tiny{$\pm$ 0.01} \\
Maximum Prob. & 0.26 \tiny{$\pm$ 0.04} & 0.18 \tiny{$\pm$ 0.05} & 0.24 \tiny{$\pm$ 0.05} & 0.13 \tiny{$\pm$ 0.02} \\
Perplexity & 0.27 \tiny{$\pm$ 0.03} & 0.17 \tiny{$\pm$ 0.04} & 0.21 \tiny{$\pm$ 0.05} & 0.11 \tiny{$\pm$ 0.03} \\
Token Entropy & 0.30 \tiny{$\pm$ 0.06} & 0.18 \tiny{$\pm$ 0.04} & 0.24 \tiny{$\pm$ 0.06} & 0.12 \tiny{$\pm$ 0.02} \\
P(True) & 0.24 \tiny{$\pm$ 0.03} & 0.21 \tiny{$\pm$ 0.06} & 0.19 \tiny{$\pm$ 0.06} & \textbf{0.17} \tiny{$\pm$ 0.02} \\
\hline
\end{tabular}

\caption{\label{tab:pr-auc} PR-AUC (considering the Not Supported class as positive) of claim-level UQ methods with FactScore labels as the ground truth (English). }
\end{table*}

  \begin{figure*}[t]
    \centering
    \includegraphics[scale=0.48,trim={0 24pt 0 0},clip]{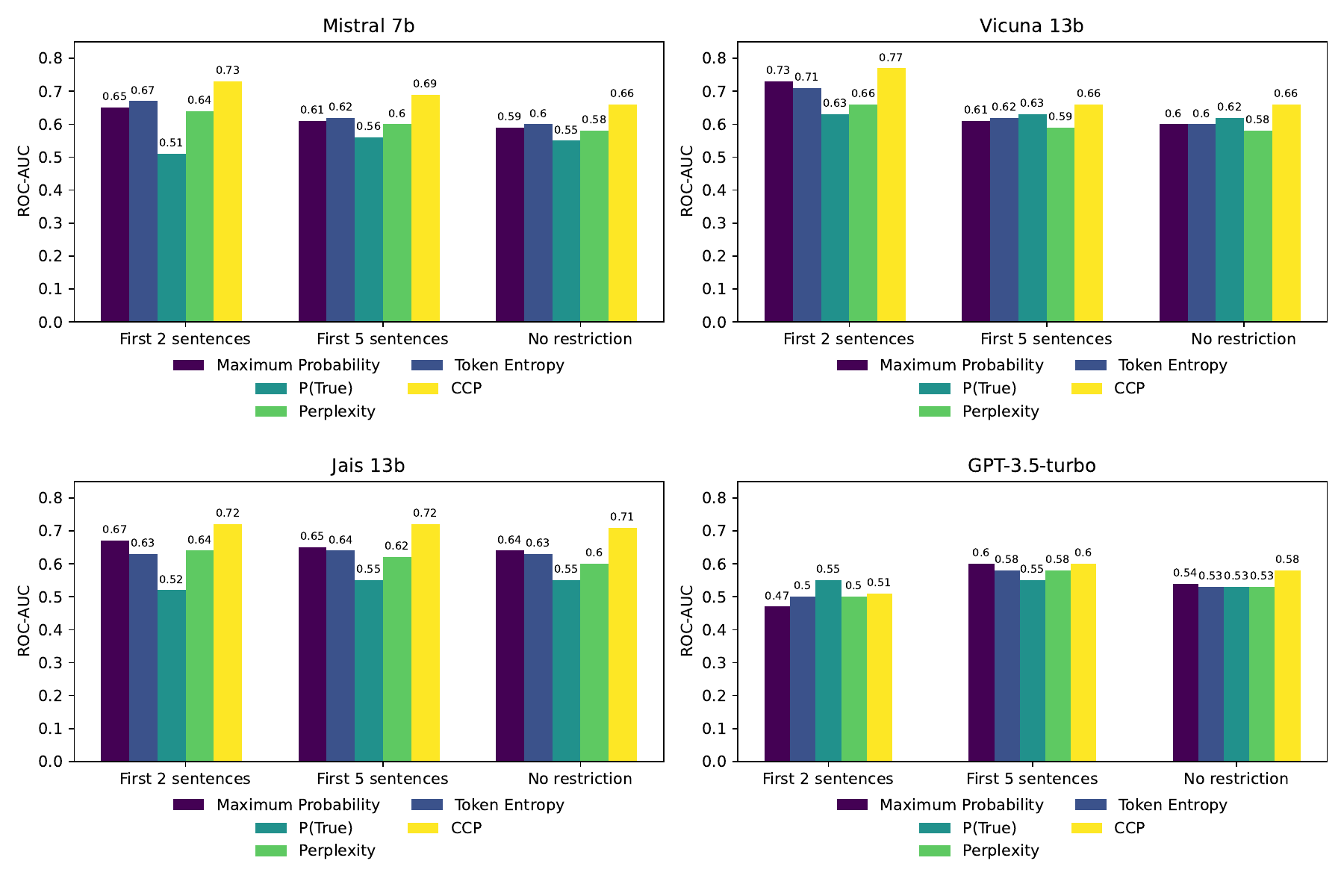}
    \caption{ROC-AUC of claim-level UQ methods based on FactScore labels, aggregated into bins when considering only facts from the first $2$, $5$, and all sentences (English).}
    \label{fig:roc-auc}
  \end{figure*}

  We decompose the generated text into atomic claims using GPT-4. For each claim, we map all its words back to generated text to access the corresponding token logits.
  Not all claims perfectly match to the original response.
  For example, for Vicuna 13b around $5\%$ of all claims do not successfully match because ChatGPT abstained to respond or outputted words not present in the original. We consider only successfully matched claims.
  
  For English, we are able to perform annotation completely automatically: each atomic claim is classified by FactScore as supported or not supported. The underlying FactScore model is ``\texttt{retrieval+ChatGPT}'' with  a dump of Wikipedia articles as an external knowledge source. We also manually annotated $100$ claims from English biographies produced by the Vicuna 13b model, $183$ claims from Arabic biographies, $146$ claims from Russian biographies, and $1603$ claims in Chinese.
  Each of the statements was checked by two annotators with access to the corresponding Wikipedia article. The final label is set to ``supported'' only if both annotators label it as so. 
 
  The statistics of the resulting datasets with automatically-labeled claims in English are presented in Table~\ref{tab:dataset-stat}, and the statistics of manually annotated datasets are presented in \Cref{tab:dataset-stat_human}. 
  The majority of claims in the model output are correct, with $6$--$29$\% of hallucinations.
  The automatic pipeline for evaluation of UQ methods using FactScore is illustrated in Figure~\ref{fig:biography_scheme_example} in Appendix~\ref{app:benchmark_details}.

\section{Experiments}
\label{sec:experiments}

\subsection{Experimental Setup}
  Fact-checking of atomic claims is framed as a binary classification task, where uncertainty scores serve as predictors of non-factuality, and FactScore or human labels serve as ground truth. The evaluation metric is ROC-AUC and PR-AUC (unsupported claims as a positive class).
  

  For the CCP method, NLI scores are calculated using the DeBERTa-large model~\cite{he2020deberta} fine-tuned for this task\footnote{\url{https://huggingface.co/microsoft/deberta-large-mnli}}. 
  The number of alternatives used in CCP is $K = 10$, except for GPT-3.5-turbo and GPT-4, as the OpenAI API does not allow us to retrieve more than $5$ alternatives from the token distribution. 
  Details on hardware and computational resources spent for experiments can be found in Appendix~\ref{sec:expenses}.

\subsection{Results for English on the FactScore Annotation}

  The main results of the experiments with FactScore labels for English are presented in Tables~\ref{tab:roc-auc} and~\ref{tab:pr-auc}. The proposed CCP method outperforms all other UQ techniques for each of the considered LLMs, with only exception in the PR-AUC metric for the GPT-3.5-turbo model, where the P(True) approach exhibits the highest performance. The underperformance of CCP for GPT-3.5-turbo may be attributed to the limited number of token options and their associated logits available through the OpenAI API. The best overall improvement from CPP is obtained for Jais 13b, where it outperforms the closest competitor by 0.07 ROC-AUC and 0.09 PR-AUC.


  We further analyze the performance by plotting ROC-AUC as a function of the number of considered sentences from the beginning of the generated text (see Figure~\ref{fig:roc-auc}). 
  We note that the quality of each method decreases as the number of considered sentences increases. This may be due to the fact that the model tends to start the response with easy-to-know and hence reliable claims and as it generates more text, it has to produce more complex and less reliable statements, which is illustrated for the Vicuna 13b model in Figure~\ref{fig:supported-frac} in the appendix.
  For the majority of cases, CCP outperforms other methods, except when we consider only the first two and the first five sentences generated by GPT-3.5-turbo.

\subsection{Multilingual Results on Manual Annotation}
\label{sec:human_eval}
  Multilingual results based on manual annotation are presented in \Cref{tab:roc-auc-human,tab:roc-auc_multilingual} and in \Cref{fig:roc-auc-multilang}.
  Using manual annotation for English, we can  also evaluate the performance of FactScore itself. The accuracy of automatic annotation is $77.2\%$ and ROC-AUC is $0.72$. A detailed analysis of FactScore mistakes is presented in Appendix~\ref{sec:factscore}.
  
  For human annotation, the performance of all UQ methods appears to be even slightly higher than for the labels obtained using FactScore (Table~\ref{tab:roc-auc-human}). 
  Moreover, CCP outperforms FactScore itself by $0.06$ ROC-AUC. These results demonstrate that in the task of detecting LLM hallucinations, UQ techniques can be a strong alternative to fact-checking tools with an external knowledge source.

  For Chinese, Arabic, and Russian, CCP also outperforms the baselines. For the Chinese Yi 6b model, from Figure~\ref{fig:roc-auc-multilang}, we can see that the gap between CCP and the baselines is especially significant for the several first claims. When considering more claims, CCP still clearly outperforms the Maximum Probability baseline, but the P(True) baseline substantially reduces the gap. For Arabic and Jais, CCP outperforms the closest competitor by $0.05$ ROC-AUC. On GPT-4 generations for Arabic, the metrics for all methods are low. We explain this observation by a small ratio of non-factual claims in the GPT-4 output. For Russian generations with the Vikhr model, CCP confidently outperforms Maximum Probability, which is its closes competitor, by $0.05$ ROC-AUC.
  

  \begin{table}
\centering
\footnotesize

\begin{tabular}{l|c|c} 
 \hline
 \diagbox[width=3.5cm]{\textbf{Method}  }{\textbf{Ground-Truth}} & \textbf{Human} & \textbf{FactScore} \\

 \hline
 CCP (ours) & \textbf{0.78} & \textbf{0.74} \\
 Maximum Prob. & 0.67 & 0.65 \\
 Perplexity & 0.65 & 0.64 \\
 Token Entropy & 0.69 & 0.65 \\
 P(True) & 0.68 & 0.65 \\
 \hline
 FactScore & 0.72 & -- \\
 \hline
\end{tabular}

\caption{\label{tab:roc-auc-human} ROC-AUC of claim-level UQ methods with human annotation and FactScore annotation as the ground truth (English, Vicuna 13b model).}
\end{table}

  \begin{table}[t]
\centering
\begin{adjustbox}{scale=0.67} 
\begin{tabular}{l|c|c|c|c}
\hline
\textbf{Model} & \textbf{\multirowcell{Yi 6b, \\ Chinese}} & \textbf{\multirowcell{Jais 13b, \\ Arabic}} & \textbf{\multirowcell{GPT-4, \\ Arabic}} & \textbf{\multirowcell{Vikhr 7b, \\ Russian}} \\
\hline
CCP (ours) & \textbf{0.64} \small{$\pm$ 0.03} & \textbf{0.66} \small{$\pm$ 0.02} & \textbf{0.56} \small{$\pm$ 0.05} & \textbf{0.68} \small{$\pm$ 0.04} \\
Maximum Prob. & 0.52 \small{$\pm$ 0.03} & 0.59 \small{$\pm$ 0.02} & 0.55 \small{$\pm$ 0.08} & 0.63 \small{$\pm$ 0.04} \\
Perplexity & 0.51 \small{$\pm$ 0.04} & 0.56 \small{$\pm$ 0.02} & 0.54 \small{$\pm$ 0.08} & 0.58 \small{$\pm$ 0.04} \\
Token Entropy & 0.57 \small{$\pm$ 0.05} & 0.61 \small{$\pm$ 0.02} & 0.48 \small{$\pm$ 0.06} & 0.55 \small{$\pm$ 0.03} \\
P(True) & 0.63 \small{$\pm$ 0.04} & 0.61 \small{$\pm$ 0.02} & 0.50 \small{$\pm$ 0.06} & 0.58 \small{$\pm$ 0.03}\\
\hline
\end{tabular}
\end{adjustbox}

\caption{\label{tab:roc-auc_multilingual} ROC-AUC of claim-level UQ methods with manual annotation as the ground truth. }
\end{table}


\subsection{Ablation Studies}
\label{sec:ablation_studies}
  In this section, we analyze the influence of various CPP components on English biographies annotated with FactScore (Tables~\ref{tab:abl-norm}--\ref{tab:abl-stopw}). Details of the experimental setup for each ablation study are presented in Appendix~\ref{app:ablation_studies}.

     \textbf{(1) Aggregation} of $CCP_{\textit{word}}$ for obtaining $CCP_{\textit{claim}}$. Besides the product of probabilities, we also tried the normalized product, minimum, and average probability. All these approaches perform slightly worse than the product (see Table~\ref{tab:abl-norm}).

     \textbf{(2) NLI model.} We investigate the influence of the specific NLI model on the performance of CCP. Table~\ref{tab:abl-nli-model} shows that CCP's effectiveness is not critically dependent on the complexity of the NLI model employed. Notably, even a relatively small model with 22M parameters  maintains strong performance without any degradation.

     \textbf{(3) NLI context.} We analyze what context is sufficient for NLI in CCP (Table~\ref{tab:abl-nli-context}). In addition to the standard variant in CCP, where we keep the claim that precedes the word in question, we experiment with a single target word without context and the whole sentence that precedes the target word. All variants demonstrate lower performance. No context results in a drop of $0.02$ ROC-AUC, and longer contexts -- of more than $0.07$.

     \textbf{(4) Functional words handling.} The results in Figure~\ref{tab:abl-stopw} show that excluding functional words in CCP helps to improve the performance by $0.03$ ROC-AUC. This approach also slightly improves the maximum probability baseline, but its performance is still much lower.

     \textbf{(5) The number of alternatives $K$.} Taking $K=5$ alternatives instead of $K=10$ in CCP decreases ROC-AUC by $0.02$. Figure~\ref{fig:nli_nsamples} also demonstrates that further decreasing $K$ reduces the performance even more. When increasing $K$, the performance plateaus at $K=8$.


\subsection{Qualitative Analysis}
  In qualitative analysis of uncertainty scores for various generations and models, we note that the maximal probability baseline produces a lot more false positives than CCP. This happens because CCP ignores some types of uncertainty, focusing only on the claim uncertainty. In some cases, CCP also finds false claims overlooked by other methods because ignoring certain types of uncertainty also allows us to reduce the cut-off threshold used to mark claims. An example where we compare CCP with maximal probability is presented in Figure~\ref{fig:token-level_comparison}, and more examples can be found in Appendix~\ref{app:examples}.

\subsection{Computational Efficiency}

  \begin{table}[t]

\centering
\footnotesize

\begin{tabular}{l|c|c}
\hline
\multirow{2}{*}{\textbf{Method}} & \textbf{NLI model} & \multirow{2}{*}{\textbf{Runtime}}\\
 & \textbf{parameters} & \\
\hline
$MP$ & --- & 18.5 \small{$\pm$ 0.8} sec \\
$CCP$ & 350M & 20.1 \small{$\pm$ 0.9} sec \\
$CCP$ & 22M & 19.1 \small{$\pm$ 0.8} sec \\
\hline
\end{tabular}

\caption{\label{tab:abl-runtime} The runtime of Maximum Probability and CCP methods on 100 biographies in English. }
\end{table}

  To demonstrate the computational efficiency of CCP, we compare it to the fastest UQ method -- Maximum Probability. Experiments were conducted using a dataset of 100 biographies and Mistral 7b~\cite{jiang2023mistral}. 
  To ensure a fair comparison, we focus solely on the runtime of generating biographies and calculating the respective uncertainty scores for each claim. Time spent on claim extraction and matching is excluded. The experiments utilized two 32GB V100 GPUs. Each biography was processed in a single batch.

  MP does not introduce notable overhead over the generation process, as it only aggregates produced logits.
  CCP involves running an NLI model for each of 10 token candidates per token position, which introduces some overhead. 

  Table~\ref{tab:abl-runtime} presents the runtime comparison. We see that NLI does not make a substantial impact. 
  Using the default microsoft/deberta-large-mnli model (350M parameters) results in 8\% increase of the runtime compared to MP. Using the smaller cross-encoder/nli-deberta-v3-xsmall model (22M parameters), which achieves comparable UQ performance, reduces the computational overhead to only 3\%.

\section{Conclusion}
\label{sec:discussion}
  We presented a novel approach to fact-checking and hallucination detection based on token-level uncertainty quantification. According to human evaluation, our approach is competitive with FactScore, a fact-checking tool that leverages an external knowledge source: we achieve similar or better results with access to only LLM outputs. 

  We proposed a computationally efficient token-level and claim-level UQ method, Claim Conditioned Probability that outperforms a number of baselines in fact-checking. In this method, we post-process the word distribution to mitigate the impact of uncertainty related to the variability of surface forms and uncertainty about what claim type to generate on the current step. In the constructed benchmark, where we detect hallucinations in biographies, CCP outperforms other methods for seven LLMs, including GPT-3.5-turbo and GPT-4, and four languages. We also demonstrate that computational overhead of CCP might be as low as 3\% of the LLM inference runtime. 

\section*{Limitations}
  While this work has been conducted according to experimental and methodological best practice, there are several potential limitations. 



  First, at the core of the approach lies a text entailment classifier. As it was originally pre-trained for a slightly different use-case, more careful analysis of its performance on diverse domains and genres should be carried out.

  Second,  the current implementation of the method makes use of OpenAI's GPT models for text segmentation and extraction of atomic facts, similarly to FactScore, which may not be practical in real applications. Replacing these components with cheaper open models should be feasible in principle, but is left for future work. 

  Third, part of our experimental results rely on a human evaluation, which may be subjective. We tried to mitigate this by creating detailed instructions for the annotators, but a larger-scale study with larger overlap would further strengthen the results.

  Fourth, our uncertainty quantification is based on tokens, where taking into account also larger units such as  noun or verb phrases as basic units of analysis may be better motivated linguistically. 

  Fifth, in this work, we do not consider the calibration of CCP scores. We note that calibration on its own does not provide the information about the performance in the practical task that we are interested in -- fact-checking. At the same time, CCP could be post-calibrated in the same way as any other probabilities or their surrogates.

  Finally, our approach only detects potentially spurious generations. A prominent direction for future research is to modify the generation of an LLM to exclude such spans, while ensuring fluency of a generated text: a simple removal of uncertain claims may result in incoherent sentences.

\section*{Ethical Considerations}
  We would like to caution that our method for fact-checking is based on uncertainty quantification, and thus it is not bullet-proof, as it only reflects the internal model state of the LLM. Thus, it is of limited utility when it comes to claims that are beyond the time cutoff of the model. Moreover, the LLM might be trained on factually false data, which is beyond our control. It could be also tricked by variations in the prompt. 

  We also caution that our solution does not eliminate hallucinations; instead, it can be used to highlight risky parts of a text, for humans to take into account. Our intended use is towards raising awareness and promoting human-machine collaboration.

  Finally, our method can be misused to unfairly moderate
  content. Thus, we ask researchers and potential users to exercise due caution.

\bibliography{ref}
\bibliographystyle{acl_natbib}

\newpage

\appendix
\clearpage
\onecolumn

\section{Fact-Checking Pipeline}

\begin{figure*}[h]
    \centering
    \includegraphics[width=\textwidth]
    {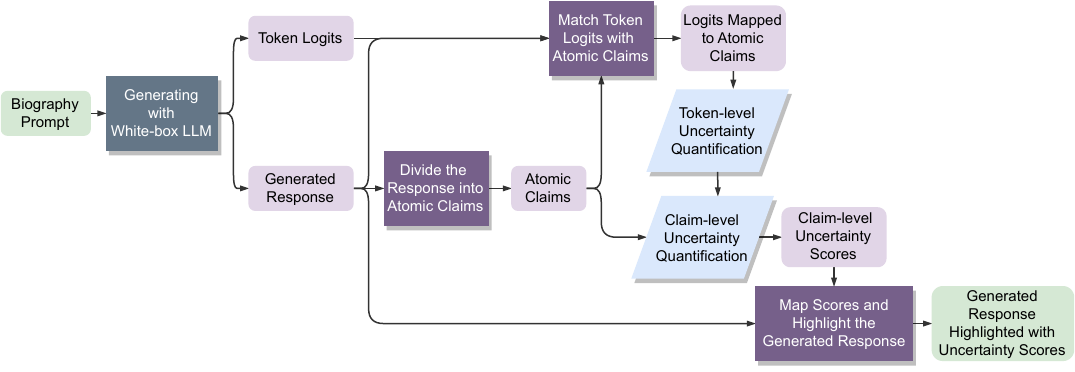}
    \caption{The scheme of the fact-checking pipeline based on UQ.}
  \label{fig:biography_scheme_general}
  \end{figure*}

\section{Example of CCP Calculation}
\label{sec:ccp_explained}
\begin{figure*}[h]
    \centering
    \includegraphics[scale=0.45]{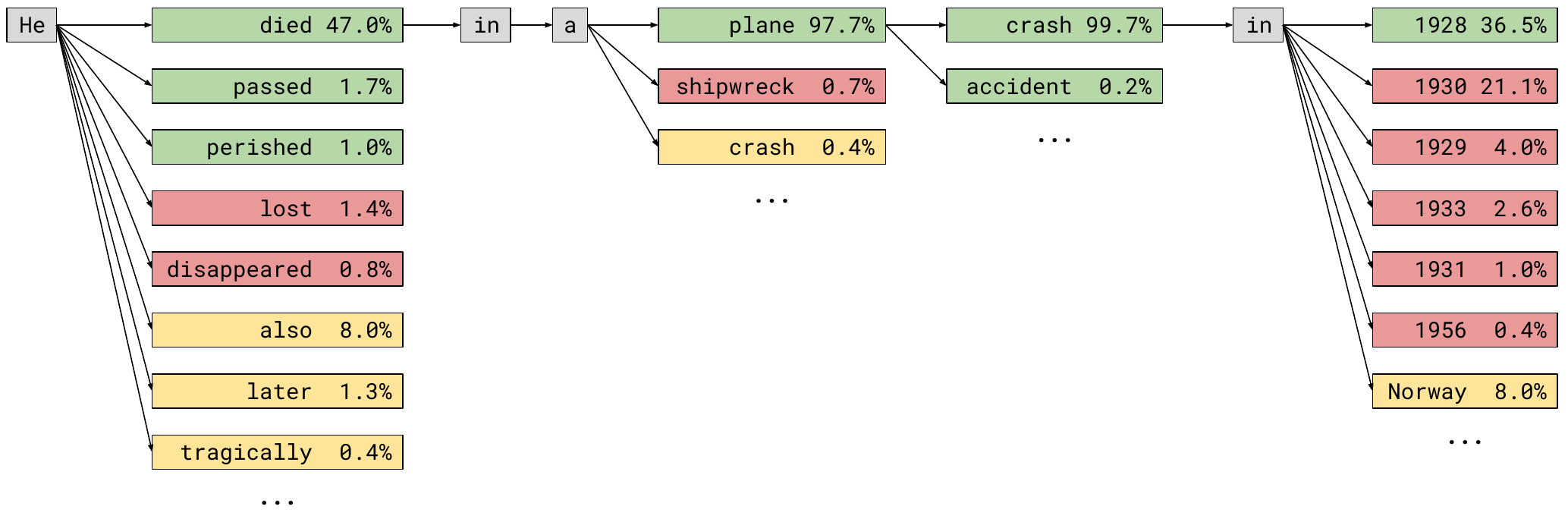}
    \caption{Example of the Vicuna 13b generation process and CCP calculation process part. The words from the greedy-generated sentence are presented sequentially on the top, each non-functional word is supplemented with its alternatives and autoregressive generation probabilities. Words with probability less than 0.1\% are omitted. \\
    Green-colored words indicate entailment to the greedy generated word, red color indicates contradiction, and yellow color indicates neutral NLI class. \\
    On the last position, CCP successfully distinguishes \texttt{Norway} from other year-related words, and does not consider its probability in the final formula. }
  \label{fig:nli_ex2}
  \end{figure*}

  Figure~\ref{fig:nli_ex2} demonstrates an example of the LLM generation process and CCP calculation process. CCP quantifies the Claim Uncertainty, not taking into account the Claim order and Surface form uncertainties. As a result, CCP produces better uncertainty scores than Maximum Probability.

\clearpage
\section{Benchmark Construction Details}
\label{app:benchmark_details}

\subsection{Benchmark Construction Pipeline}

\begin{figure*}[h]
    \centering
    \includegraphics[width=\textwidth]{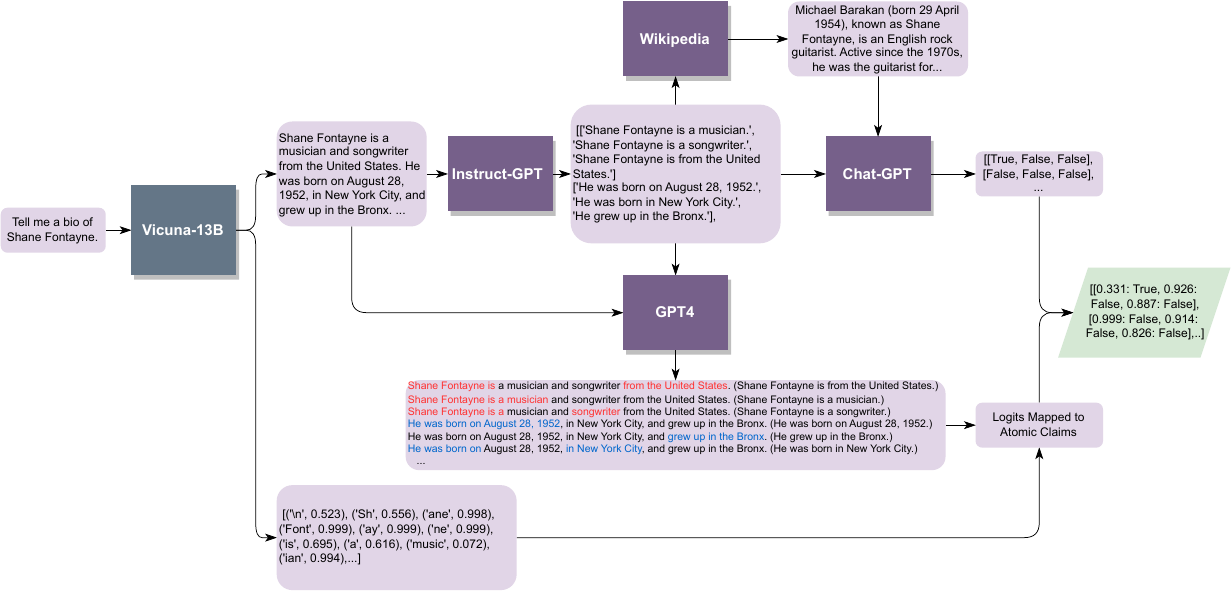}
    \caption{A visualization of the benchmark construction pipeline based on FactScore with an example.}
  \label{fig:biography_scheme_example}
  \end{figure*}

  Figure~\ref{fig:biography_scheme_general} presents the suggested general preparation pipeline of a factuality dataset for an arbitrary model and Fact-Checking benchmark.
  This pipeline was used to generate the biography dataset. 
  Figure~\ref{fig:biography_scheme_example} presents the detailed pipeline of the biography dataset preparation, based on the general schema.
  
  The prompt for the matching model to map atomic claims to the initial text is as follows: ``\texttt{Given the fact ``{fact}'', identify the corresponding words in the original sentence ``{sent}'' that help derive this fact. Please list all words that are related to the fact, in the order they appear in the original sentence, each word separated by comma.}''.
  The prompt for the model that partitions the initial generated text into individual atomic claims, as well as the prompt for the classification model, are taken similar to the paper by~\citet{min2023factscore}.

\subsection{Datasets and Statistics}

\begin{table}[h]
\centering
\footnotesize

\begin{tabular}{l|c|c} 
 \hline
 \textbf{Model}      & \textbf{\multirowcell{Number of \\ claims}} & \textbf{\multirowcell{Supported \\ claims}} \\
 \hline
 Mistral 7b    & 3,824 & 71.1\% \\
 Vicuna 13b    & 3,617 & 78.2\% \\ 
 Jais 13b      & 1,407 & 84.3\% \\
 GPT-3.5-turbo & 3,875 & 89.4\% \\
 \hline
 \end{tabular}


\caption{\label{tab:dataset-stat} The statistics of the datasets generated from 100 biographies using all tested English LLMs and annotated \underline{automatically}.}
~~~~~~
\centering
\footnotesize

\begin{tabular}{l|c|c} 
 \hline
 \textbf{Model}      & \textbf{\multirowcell{Number of \\ claims}} & \textbf{\multirowcell{Supported \\ claims}} \\
 \hline
 Vicuna 13b, English & 100 & 70.1\% \\
 Yi 6b, Chinese & 1,603 & 94.0\% \\
 Jais 13b, Arabic & 186 & 73.0\% \\
 GPT-4, Arabic & 200 & 92.5\% \\
 Vikhr 7b, Russian & 146 & 72.6\% \\
 \hline
 \end{tabular}

\caption{\label{tab:dataset-stat_human} The statistics of the datasets generated from 100 biographies using all tested English LLMs and annotated \underline{manually}.}
\end{table}

  Since FactScore only supports English, for Arabic, Chinese, and Russian, we generate biographies of well-known people and annotate them only manually. We also manually annotate 100 English claims generated by Vicuna 13b. The statistics for the annotated datasets are presented in Table~\ref{tab:dataset-stat_human}.

  For Arabic, using GPT-4, we generate 100 biographies of people randomly selected from the list of the most visited websites in Arabic Wikipedia.
  The used Arabic prompt is the translation of: ``Tell me the biography of \{person name\}''.
  To extract claims, we prompt GPT-4 in the following way: ``Convert the following biography into Arabic atomic factual claims that can be verified, one claim per line.
  Biography is: \{biography\}''.
  Arabic biographies and claims are translated into English using Google Translate.
  It is worth mentioning that almost one-third of the names in the list of person names are foreign, e.g.  Donald Trump, Messi, Isaac Newton, Pope Benedict XVI, etc.
  On average, GPT-4 generates 20 claims from each biography, and random two claims from each biography are verified manually (total = 200 claims). 
  
  For Jais 13b experiments, we use the same prompts used for GPT-4. We notice that the biographies generated by Jais 13b are much shorter than the ones generated by GPT-4 (almost half-length). Similarly, we use GPT-4 to extract claims from the generated biographies. On average, biographies generated by Jais 13b have nine claims. Jais 13b generates empty biographies for seven names (out of 100) with response messages like: ``I am sorry! I cannot provide information about \{name\}'', or ``What do you want to know exactly?''. Two random claims from each biography are verified manually (total = 186 claims).
  
  For Chinese, we first prompt ChatGPT to generate a list of 100 famous people. Then use the same way as we have done in Arabic, but change the prompt to Chinese, to generate biographies and claims. We use Yi 6b to generate texts and GPT-4 to split them into atomic claims.

  For Russian, we conduct a similar approach, prompting ChatGPT to generate a list of 100 famous people and checking the result to obtain representative personalities from different areas such as science, sport, literature, art, government activity, cinematography, heroes, etc. A balanced list of famous people in different professional categories was obtained. For these people, we generate biographies using the Vikhr 7b model \cite{nikolich2024vikhr}.

\subsection{FactScore Annotation}
\label{sec:factscore}
  The statistics of English data annotated using FactScore is presented in Table \ref{tab:dataset-stat}. Here we give examples of the operation of the FactScore automatic markup system, see the Table~\ref{tab:factscore-fails}.
  We hypothesize that the causes of FactScore errors are related to model hallucination (true information is present in the knowledge source, but the model produces incorrect information), lack of context (an excerpt from a Wikipedia article cannot capture all the details), and difficulties with information interpretation (the desired information is present in the knowledge base, but is formulated in different words or in several sentences).
  All these cases leave room for further improvement of the pipeline.
  We have also given the results of the proposed CCP method on selected examples. It can be seen that the results of our algorithm are similar to the annotation from FactScore, while CCP does not use external information in calculating the scores.

  \begin{table*}[h]
\small
\centering
\resizebox{\textwidth}{!}{%
\begin{tabular}{l|l|l|l|p{5cm}|p{9cm}}
\toprule
\textbf{Cases} & \textbf{FS}    & \textbf{Human} & \textbf{CCP} & \textbf{Generated Atomic Claim}                                                                & \textbf{True Information from the Wikipedia article}                                                                                                     \\
\hline
\hline
TN    & False & False & 0.819 & Marie Stopes died on October 20, 1958. & Marie Charlotte Carmichael Stopes (15 October 1880 – 2 October 1958) was a British author...\\
\hline
TN    & False & False & 0.999 & Planck is best known for his work on the nature of light.& His fame as a physicist rests primarily on his role as the originator of quantum theory...\\
\hline
\hline
FN    & False & True & 1.0 & Heisenberg was appointed as the director of the Max Planck Institute. & He then became director of the Max Planck Institute for Physics and Astrophysics from 1960 to 1970.\\
\hline
FN    & False & True & 0.716 & Ray Charles incorporated elements of Latin music into his sound.& Charles reached the pinnacle of his success at Atlantic with the release of "What'd I Say", which combined gospel, jazz, blues and Latin music. \\
\hline
\hline
FP    & True  & False & 0.001 & Sagan was a prolific writer.& Carl Edward Sagan (November 9, 1934 – December 20, 1996) was an American astronomer and science communicator.    \\
\hline
FP    & True  & False & 0.141 & Van Gogh was a pastor.& Van Gogh prepared for the University of Amsterdam theology entrance examination; he failed the exam ...\\
\hline
FP    & True  & False & 0.185 & Hawking showed an early aptitude for science.& Although known at school as "Einstein", Hawking was not initially successful academically.\\
\hline
\hline
TP    & True  & True &  0.276 & Tiger Woods is a professional golfer.& Eldrick Tont "Tiger" Woods (born December 30, 1975) is an American professional golfer.\\
\hline
TP    & True  & True & 0.015 & Miles Davis began playing the trumpet at the age of 13.& On his thirteenth birthday his father bought him a new trumpet,[17] and Davis began to play in local bands.\\
\hline
TP    & True  & True & 0.001 & Hitchcock directed "The Birds."& Hitchcock's other notable films include Rope (1948), Strangers on a Train (1951), ..., Birds (1963) and  Marnie (1964), ...   \\   
\hline
\bottomrule
\end{tabular}
}
\caption{\label{tab:factscore-fails} Table with examples of the FactScore automatic annotation system and all types of classification outcomes when comparing automatic annotation and manual annotation (confusion matrix elements): True Negative (TN), False Negative (FN), False Positive (FP), True Negative (TN); FS is a FactScore label, Human is a human annotation label. CCP scores are comparable to FactScore annotation and do not require an external source of information.}
\end{table*}

\section{Ablation Studies}
\label{app:ablation_studies}
  In this section, we perform the ablation study of various CCP components.

\subsection{Aggregation}
\begin{table}[h]

\centering
\footnotesize

\begin{tabular}{l|c|c}
\hline
\textbf{Method} & \textbf{ROC-AUC} & \textbf{PR-AUC} \\
\hline
$CCP_\textit{prod}$ & \textbf{0.66} \small{$\pm$ 0.03} & \textbf{0.22} \small{$\pm$ 0.05} \\
$CCP_\textit{len}$ & 0.65 \small{$\pm$ 0.03} & 0.21 \small{$\pm$ 0.03} \\
$CCP_\textit{min}$ & 0.64 \small{$\pm$ 0.03} & 0.13 \small{$\pm$ 0.04} \\
$CCP_\textit{mean}$ & 0.65 \small{$\pm$ 0.03} & \textbf{0.22} \small{$\pm$ 0.04} \\
\hline
\end{tabular}

\caption{\label{tab:abl-norm} ROC-AUC and PR-AUC on Vicuna 13b generation, for different normalizations. }
\end{table}

  Table~\ref{tab:abl-norm} presents the result of CCP on Vicuna 13b when applying $4$ different kinds of aggregation:
  \begin{align*}
    CCP_{\textit{prod}}(F) &= 1 - \prod_{j=1}^k CCP_{\textit{word}}(x_j),\\
    CCP_{\textit{len}}(F) &= 1 - \exp\biggl(\frac{1}{k}\sum_{j=1}^k \log CCP_{\textit{word}}(x_j)\biggr),\\
    CCP_{\textit{min}}(F) &= 1 - \min_{j = 1, \dots, k} CCP_{\textit{word}}(x_j), \\
    CCP_{\textit{mean}}(F) &= 1 - \frac{1}{k}\sum_{j=1}^k CCP_{\textit{word}}(x_j).
  \end{align*}
  $CCP_{\textit{prod}}$ shows the best results and is chosen for the final $CCP$ formula.

\subsection{NLI Models}

  \begin{table}

\centering
\footnotesize

    \begin{tabular}{l|l|c|c}
        \hline
        \textbf{NLI model} & \textbf{Parameters} & \textbf{ROC-AUC} & \textbf{PR-AUC} \\
        \hline
        microsoft/deberta-large-mnli & 350M & \textbf{0.66} \small{$\pm$ 0.06} & \textbf{0.24} \small{$\pm$ 0.04} \\
        microsoft/deberta-base-mnli & 86M & 0.65 \small{$\pm$ 0.06} & 0.23 \small{$\pm$ 0.05} \\
        MoritzLaurer/mDeBERTa-v3-base & 86M & \textbf{0.66} \small{$\pm$ 0.06} & 0.21 \small{$\pm$ 0.03} \\
        cross-encoder/nli-deberta-v3-large & 350M & \textbf{0.66} \small{$\pm$ 0.06} & 0.22 \small{$\pm$ 0.04} \\
        cross-encoder/nli-deberta-v3-base & 86M & 0.65 \small{$\pm$ 0.06} & 0.23 \small{$\pm$ 0.05} \\
        cross-encoder/nli-deberta-v3-small & 44M & 0.65 \small{$\pm$ 0.06} & 0.22 \small{$\pm$ 0.05} \\
        cross-encoder/nli-deberta-v3-xsmall & 22M & \textbf{0.66} \small{$\pm$ 0.07} & \textbf{0.24} \small{$\pm$ 0.06} \\
        \hline
    \end{tabular}

    \caption{ The ROC-AUC and PR-AUC metrics of CCP on biographies generation dataset with Vicuna 13b model, when using different NLI models with specified identifiers in Huggingface library.}
    
    \label{tab:abl-nli-model}
\end{table}

  Table~\ref{tab:abl-nli-model} presents the result of CCP on Vicuna 13b when using different NLI models. We selected fine-tuned NLI models from HuggingFace, encompassing a range of sizes from Microsoft's DeBERTa model~\cite{he2020deberta}, a multilingual variant~\cite{laurerless2022}, and a model from SentenceTransformers~\cite{reimers-2019-sentence-bert}.

  Our findings demonstrate that the CCP performance exhibits minimal dependence on the specific NLI model chosen. Notably, even CCP utilizing a relatively small CrossEncoder model with only 22M parameters achieves the highest performance.

\subsection{NLI Context}

  \begin{table}[t!]

\centering
\footnotesize

\begin{tabular}{l|c|c}
\hline
\textbf{Method} & \textbf{ROC-AUC} & \textbf{PR-AUC} \\
\hline
$CCP_\textit{no context}$ & 0.64 \small{$\pm$ 0.03} & \textbf{0.24} \small{$\pm$ 0.04} \\
$CCP_\textit{sent pref}$ & 0.59 \small{$\pm$ 0.03} & 0.18 \small{$\pm$ 0.04}  \\
$CCP_\textit{claim pref}$ & \textbf{0.66} \small{$\pm$ 0.03} & \textbf{0.24} \small{$\pm$ 0.05}  \\
\hline
\end{tabular}

\caption{\label{tab:abl-nli-context} ROC-AUC and PR-AUC on Vicuna 13b generation, for different contexts to input words with to an NLI model. }
\end{table}

  Table~\ref{tab:abl-nli-context} presents the result of CCP on Vicuna 13b when adding different contexts as the inputs of NLI model:

  \begin{enumerate}
    \item $CCP_\textit{no context}$ calculates $NLI$ using these $2$ words as NLI model input;

    \item $CCP_\textit{sent pref}$ calculates $NLI$ using the prefix in the sentence from model generation up to the current word and its alternative, as input to NLI model;

    \item $CCP_\textit{claim pref}$, uses all words in the sentence corresponding to the greedy word, which were matched to the current fact and which comes before the greedy word (see example on Figure~\ref{fig:nli_ex1}).
  \end{enumerate}

  The results for sentence prefix provides too wide context to the NLI model, which results in the NLI model focusing on the contexts instead of the greedy words and its alternatives, thus outputting lots of \text{entailment} classes. On the other hand, the method without context does not provide enough context to calculate NLI class with more quality. $CCP_{\textit{claim pref}}$ shows better results and is the one chosen for the final $CCP$ formula.

\subsection{Functional Words Handling}

  \begin{table}[h]

\centering
\footnotesize

\begin{tabular}{l|c|c}
\hline
\textbf{Method} & \textbf{ROC-AUC} & \textbf{PR-AUC}\\
\hline
$CCP_\textit{confident}$ & \textbf{0.66} \small{$\pm$ 0.03} & \textbf{0.24} \small{$\pm$ 0.05} \\
$CCP_\textit{ignore}$ & 0.63 \small{$\pm$ 0.03} & 0.23 \small{$\pm$ 0.06} \\
$MP_\textit{confident}$ & 0.59 \small{$\pm$ 0.05} & 0.18 \small{$\pm$ 0.04} \\
$MP_\textit{ignore}$ & 0.60 \small{$\pm$ 0.05} & 0.19 \small{$\pm$ 0.05} \\
\hline
\end{tabular}

\caption{\label{tab:abl-stopw} ROC-AUC and PR-AUC on Vicuna 13b generation, for methods of handling functional words. }
\end{table}

  Table~\ref{tab:abl-stopw} presents the result of CCP and Maximum Probability on Vicuna 13b when handling functional words differently. The 2 approaches tested are:

  \begin{enumerate}
    \item $CCP_{confident}$ and $MP_{confident}$ are the versions of $CCP$ and $MP$ baselines, which assigns the most confident word-level score of 1.0 to words from NLTK~\cite{nltk} stopwords list, which is the same as skipping these words form the list of matched words of length $k$;

    \item $CCP_{ignore}$ and $MP_{ignore}$ handles functional words similar to any other words. $MP_{ignore}$ is the baseline used in the main section.
  \end{enumerate}

  $CCP_{\textit{confident}}$ shows better results and is the one chosen for final $CCP$ formula. In contract, Maximum Probability performs worse if specifically handling functional words.

\subsection{Number of Alternatives}
\begin{figure}[h]
    \centering
    \includegraphics[scale=0.5,trim={1cm 0cm 1.5cm 0.5cm}]{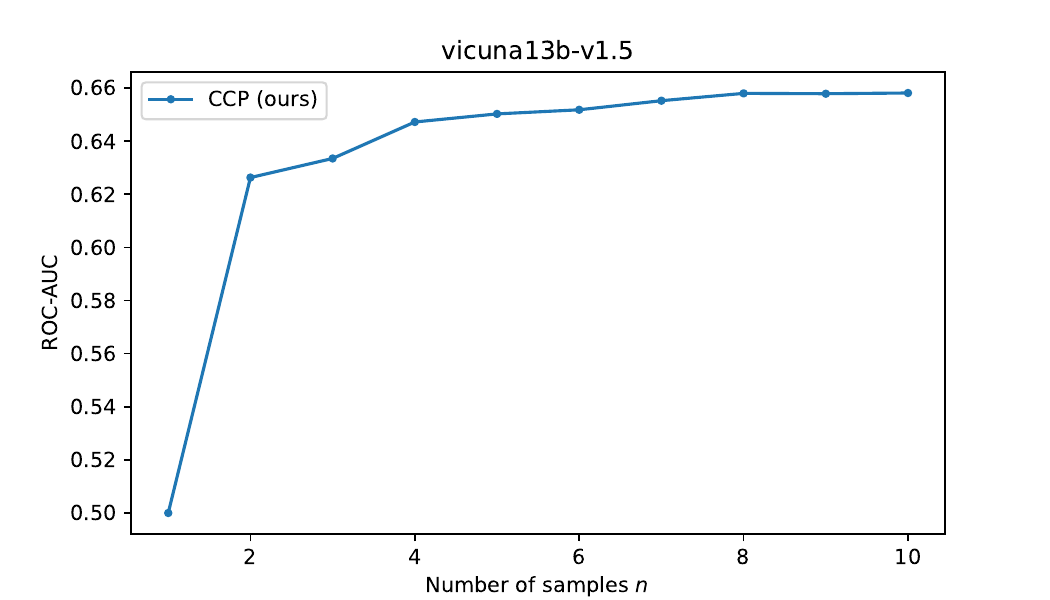}
    \caption{ROC-AUC between FactScore classes and the claim-level CCP method as a function of $n$ -- the number of token options in the probability distribution of the LLM to consider.}
  \label{fig:nli_nsamples}
  \end{figure}

  Figure~\ref{fig:nli_nsamples} presents the CCP ROC-AUC results on the whole model generation, depending on the number of beams $n$ to run the method with.

\section{Additional Experimental Results}
\begin{figure}[h]
    \centering
    \includegraphics[scale=0.5,trim={1cm 0cm 1.5cm 0.5cm}]{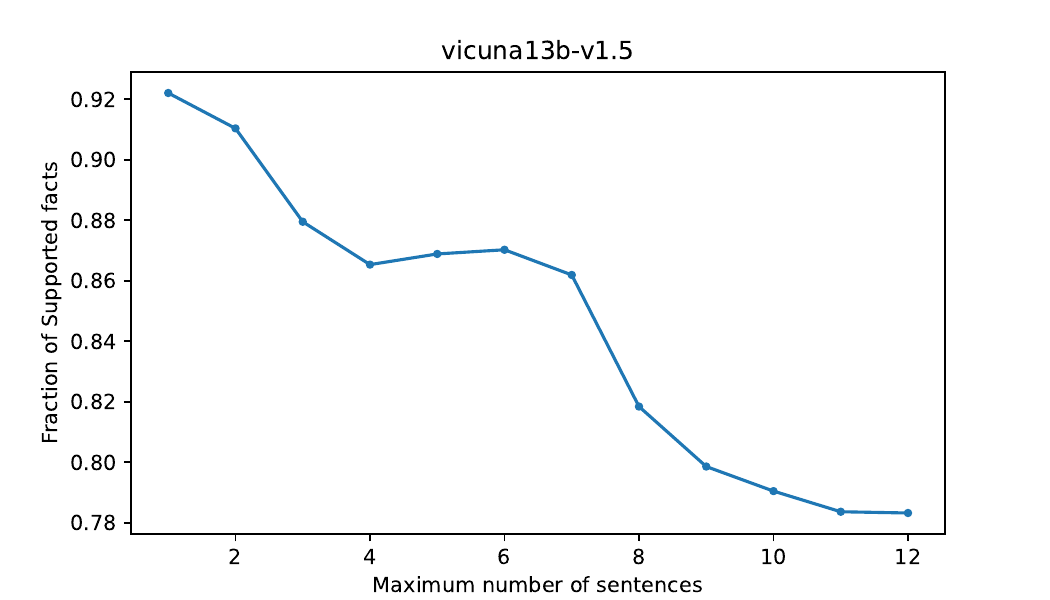}
    \caption{Percentage of supported claims, as a function of the number of sentences to restrict generation to (English, Vicuna 13b).}
  \label{fig:supported-frac}
  \end{figure}

  Here we show additional results related to the performance of our pipeline and CCP method.
  In the Figure~\ref{fig:supported-frac} we show the dependence of the percentage of supported claims in the generated LLM response (in this case for Vicuna 13b) as a function of its length. 
  As we can see, the LLM produces wrong claims when the generation length increases, which may be due to the generation of additional facts.

\begin{figure}[H]
    \centering
    \includegraphics[scale=0.5]{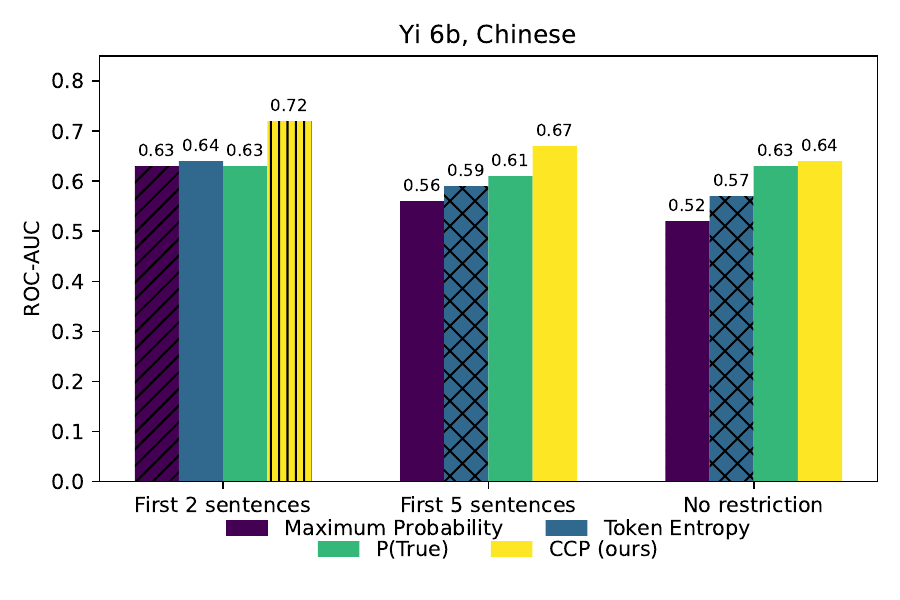}
    \caption{The comparison of token-level uncertainty quantification methods in terms of ROC-AUC scores, measured for Chinese dataset.
    The results are split into bins when considering only facts from the first $2$, $5$, and all sentences.}
  \label{fig:roc-auc-multilang}
  \end{figure}

  In Figure~\ref{fig:roc-auc-multilang}, we show the performance quality of our CCP method on Chinese data on the Yi 6b model.
  We see that, as for other English models, our method outperforms the alternatives over the entire generation length in terms of ROC-AUC.





\section{Resources and Expenses}
\label{sec:expenses}
  For a single run of data generation and UQ methods evaluation, we spent 12 days of Nvidia A100 GPU compute time. OpenAI API was used mainly for splitting and matching atomic claims, the total cost for all English models (Mistral 7b, Vicuna 13b, Jais 13b, GPT-3.5-turbo) was \$40. The cost of generating English biographies with GPT-3.5-turbo was \$13.

\newpage

\section{Examples of Results Obtained with CCP and the Maximal Probability Baseline}
\label{app:examples}

\begin{figure*}[h]
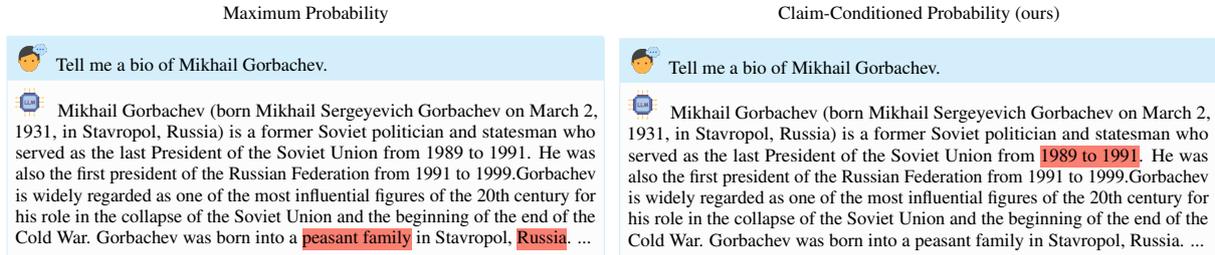
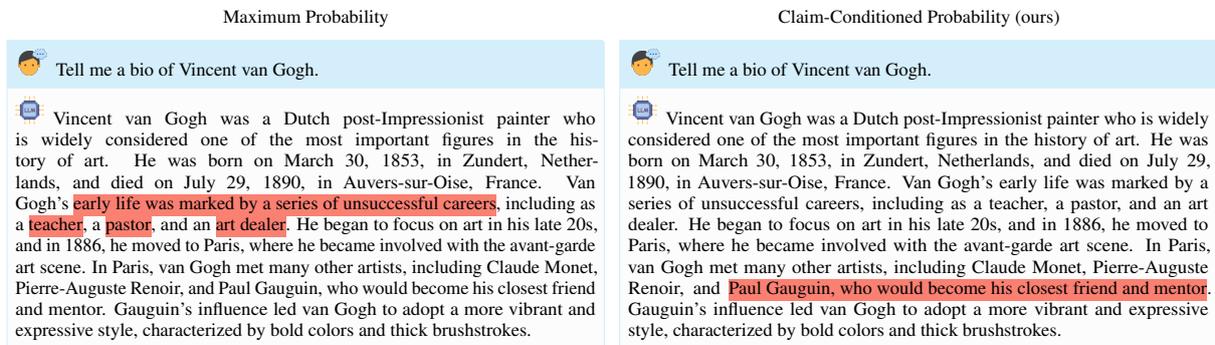
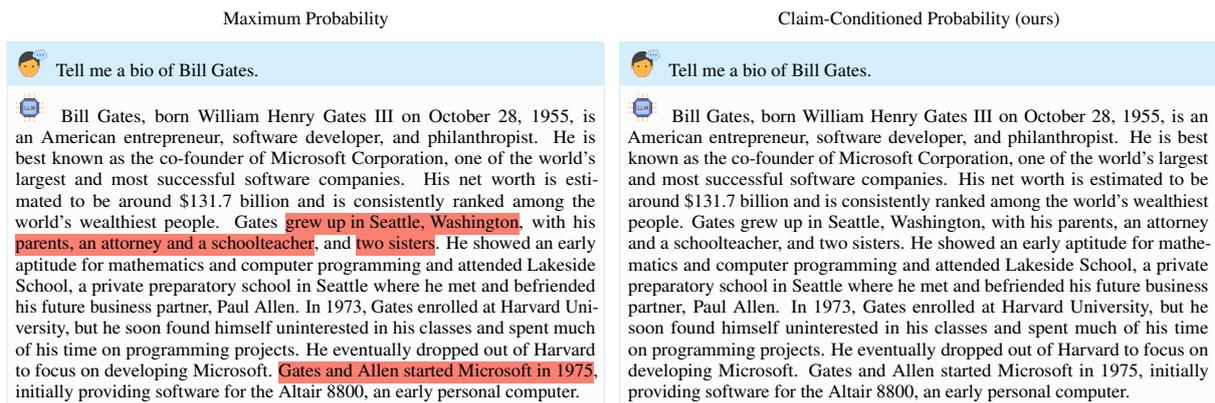

	\centering
	\scriptsize
 \begin{multicols}{2}

Maximum Probability
 
	\begin{tcolorbox}[size=fbox,
		title=\,{\includegraphics[width=1.5em]{tex/img/user-icon.png}\,\,\,Tell me a bio of Mikhail Gorbachev.},
		colback=gray!2!white,
		colframe=cyan!15!white,
            coltitle=black,
		width=32em,
		sharp corners=northwest,
		]
			\includegraphics[width=1.5em]{tex/img/llm-icon.png}
			 \,\,\,Mikhail Gorbachev (born Mikhail Sergeyevich Gorbachev on March 2, 1931, in Stavropol, Russia) is a former Soviet politician and statesman who served as the last President of the Soviet Union from 1989 to 1991. He was also the first president of the Russian Federation from 1991 to 1999.Gorbachev is widely regarded as one of the most influential figures of the 20th century for his role in the collapse of the Soviet Union and the beginning of the end of the Cold War.
Gorbachev was born into a \marktext{myorange}{peasant family} in Stavropol, \marktext{myorange}{Russia}.  ...

	\end{tcolorbox}

\columnbreak

Claim-Conditioned Probability (ours)

\begin{tcolorbox}[size=fbox,
		title=\,{\includegraphics[width=1.5em]{tex/img/user-icon.png}\,\,\,Tell me a bio of Mikhail Gorbachev.},
		colback=gray!2!white,
		colframe=cyan!15!white,
            coltitle=black,
		width=32em,
		sharp corners=northwest,
		]
			\includegraphics[width=1.5em]{tex/img/llm-icon.png}
			\,\,\,Mikhail Gorbachev (born Mikhail Sergeyevich Gorbachev on March 2, 1931, in Stavropol, Russia) is a former Soviet politician and statesman who served as the last President of the Soviet Union from \marktext{myorange}{1989 to 1991}. He was also the first president of the Russian Federation from 1991 to 1999.Gorbachev is widely regarded as one of the most influential figures of the 20th century for his role in the collapse of the Soviet Union and the beginning of the end of the Cold War.
Gorbachev was born into a peasant family in Stavropol, Russia.  ...
	\end{tcolorbox}
 
 \end{multicols}
	    \caption{The Maximum Probability baseline vs Claim-Conditioned Probability (ours).}
	\label{fig:token-level_comparison_app2}
\end{figure*}

\begin{figure*}[h]
	\centering
	\scriptsize
 \begin{multicols}{2}

Maximum Probability
 
	\begin{tcolorbox}[size=fbox,
		title=\,{\includegraphics[width=1.5em]{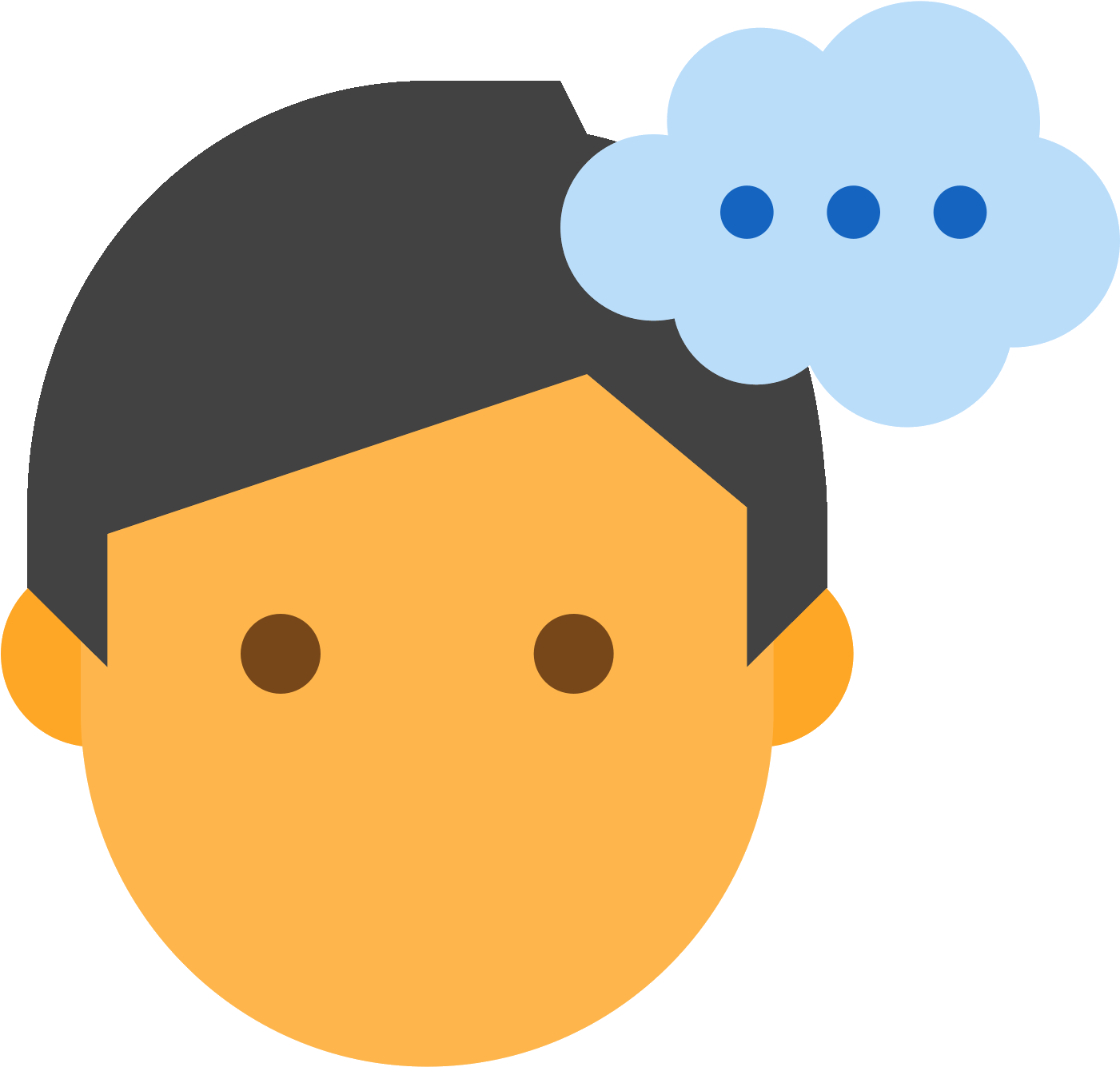}\,\,\,Tell me a bio of Vincent van Gogh.},
		colback=gray!2!white,
		colframe=cyan!15!white,
            coltitle=black,
		width=32em,
		sharp corners=northwest,
		]
			\includegraphics[width=1.5em]{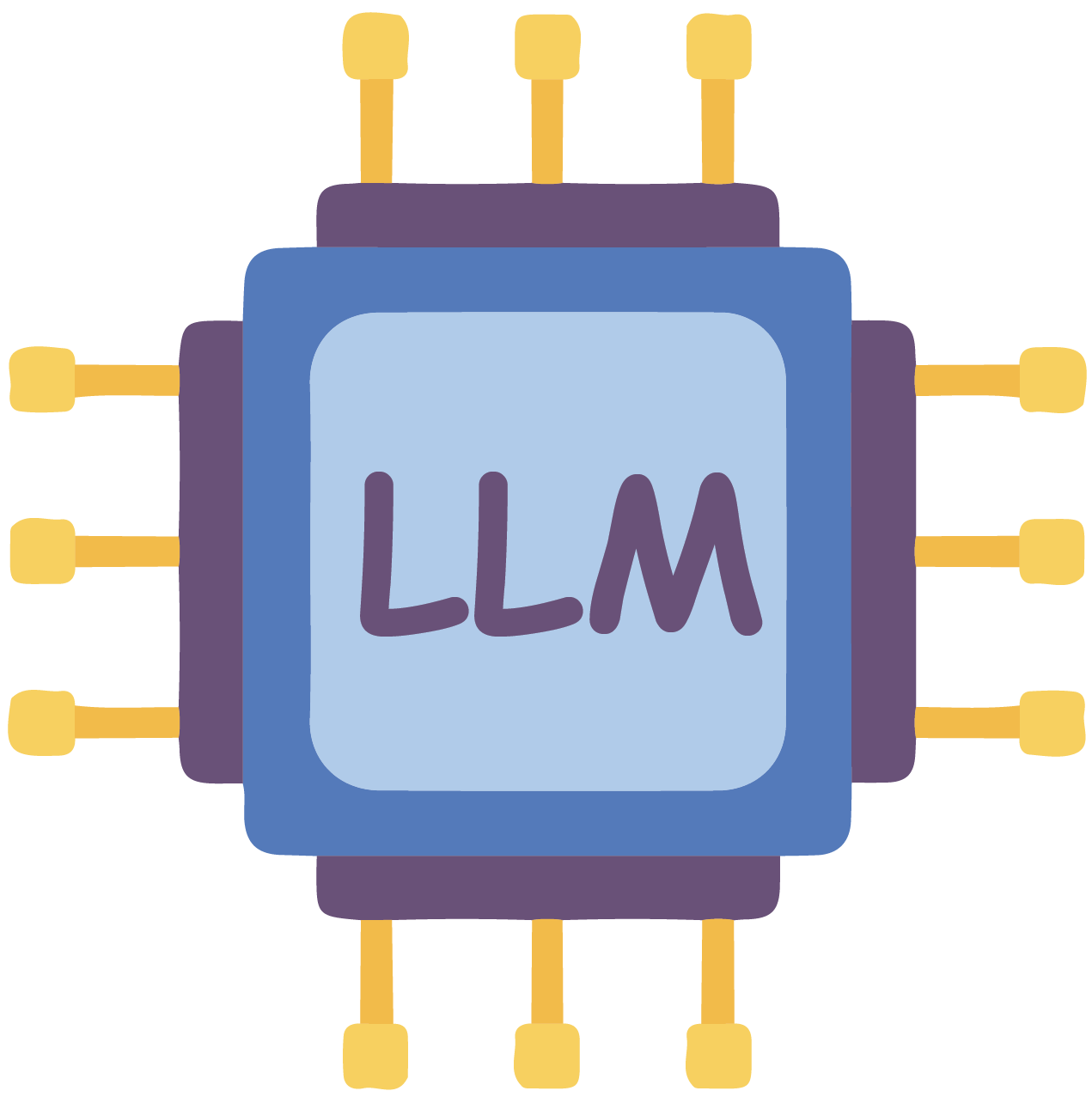}\,\,\,Vincent van Gogh was a Dutch post-Impressionist painter who is widely considered one of the most important figures in the history of art. He was born on March 30, 1853, in Zundert, Netherlands, and died on July 29, 1890, in Auvers-sur-Oise, France.
Van Gogh's \marktext{myorange}{early life was marked by a series of unsuccessful careers}, including as a \marktext{myorange}{teacher}, a \marktext{myorange}{pastor}, and an \marktext{myorange}{art dealer}. He began to focus on art in his late 20s, and in 1886, he moved to Paris, where he became involved with the avant-garde art scene.
In Paris, van Gogh met many other artists, including Claude Monet, Pierre-Auguste Renoir, and Paul Gauguin, who would become his closest friend and mentor. Gauguin's influence led van Gogh to adopt a more vibrant and expressive style, characterized by bold colors and thick brushstrokes.

	\end{tcolorbox}

\columnbreak

Claim-Conditioned Probability (ours)

\begin{tcolorbox}[size=fbox,
		title=\,{\includegraphics[width=1.5em]{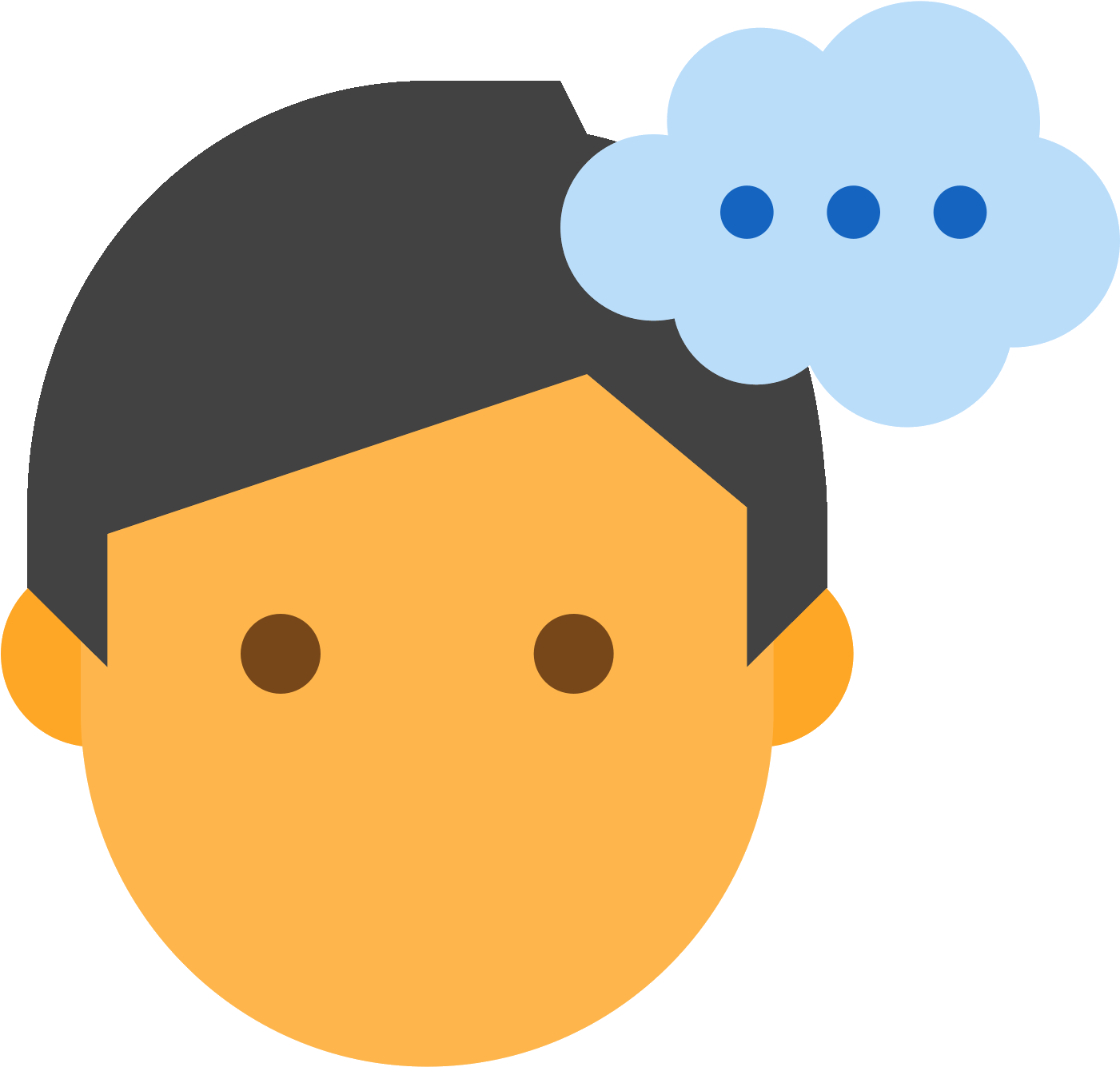}\,\,\,Tell me a bio of Vincent van Gogh.},
		colback=gray!2!white,
		colframe=cyan!15!white,
            coltitle=black,
		width=32em,
		sharp corners=northwest,
		]
			\includegraphics[width=1.5em]{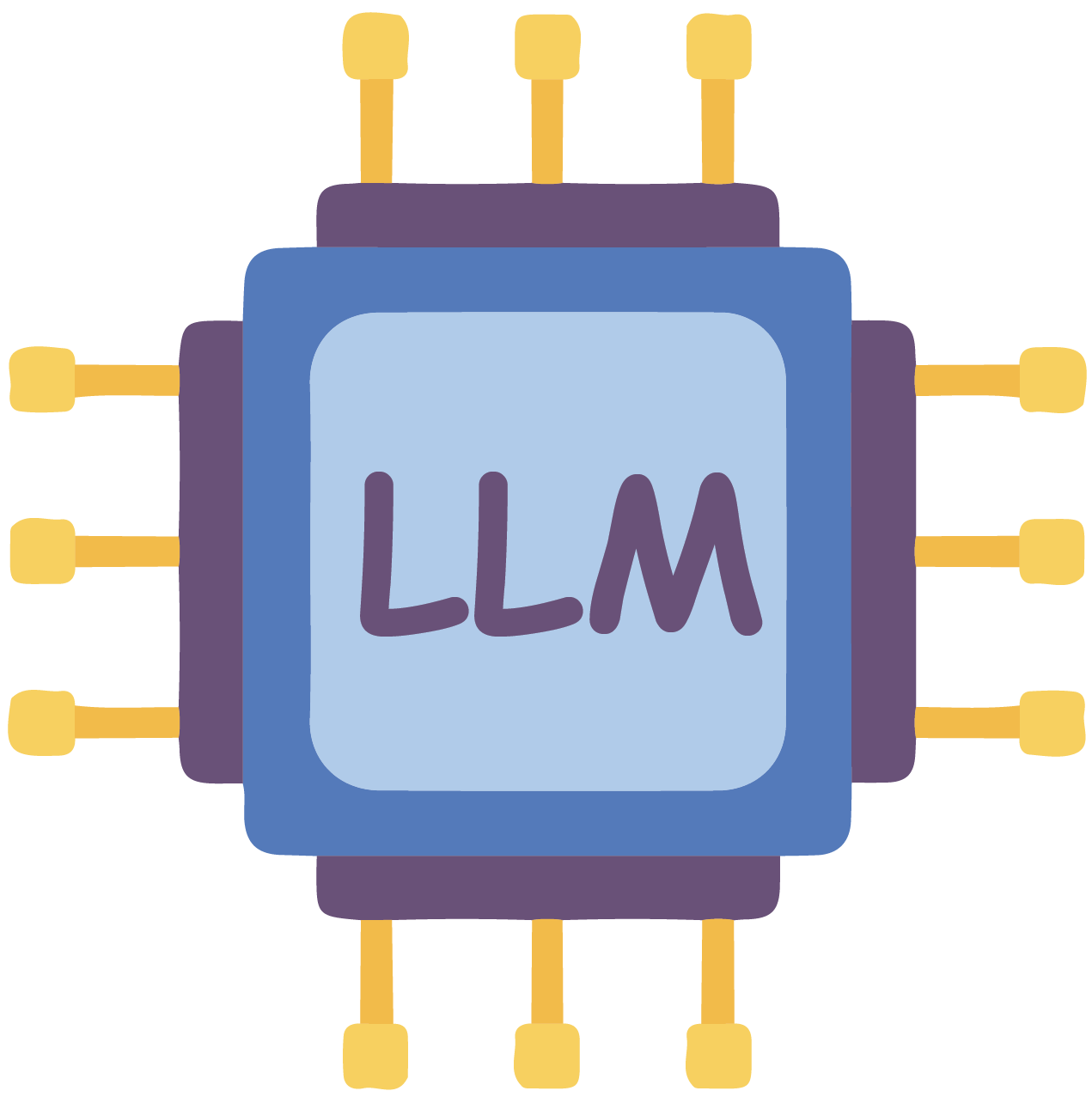}\,\,\,Vincent van Gogh was a Dutch post-Impressionist painter who is widely considered one of the most important figures in the history of art. He was born on March 30, 1853, in Zundert, Netherlands, and died on July 29, 1890, in Auvers-sur-Oise, France.
Van Gogh's early life was marked by a series of unsuccessful careers, including as a teacher, a pastor, and an art dealer. He began to focus on art in his late 20s, and in 1886, he moved to Paris, where he became involved with the avant-garde art scene.
In Paris, van Gogh met many other artists, including Claude Monet, Pierre-Auguste Renoir, and \marktext{myorange}{Paul Gauguin, who would become his closest friend and mentor}. Gauguin's influence led van Gogh to adopt a more vibrant and expressive style, characterized by bold colors and thick brushstrokes.

	\end{tcolorbox}
 
 \end{multicols}
	    \caption{The Maximum Probability baseline vs Claim-Conditioned Probability (ours).}
	\label{fig:token-level_comparison_app}
\end{figure*}

\begin{figure*}[h]
	\centering
	\scriptsize
 \begin{multicols}{2}

Maximum Probability
 
	\begin{tcolorbox}[size=fbox,
		title=\,{\includegraphics[width=1.5em]{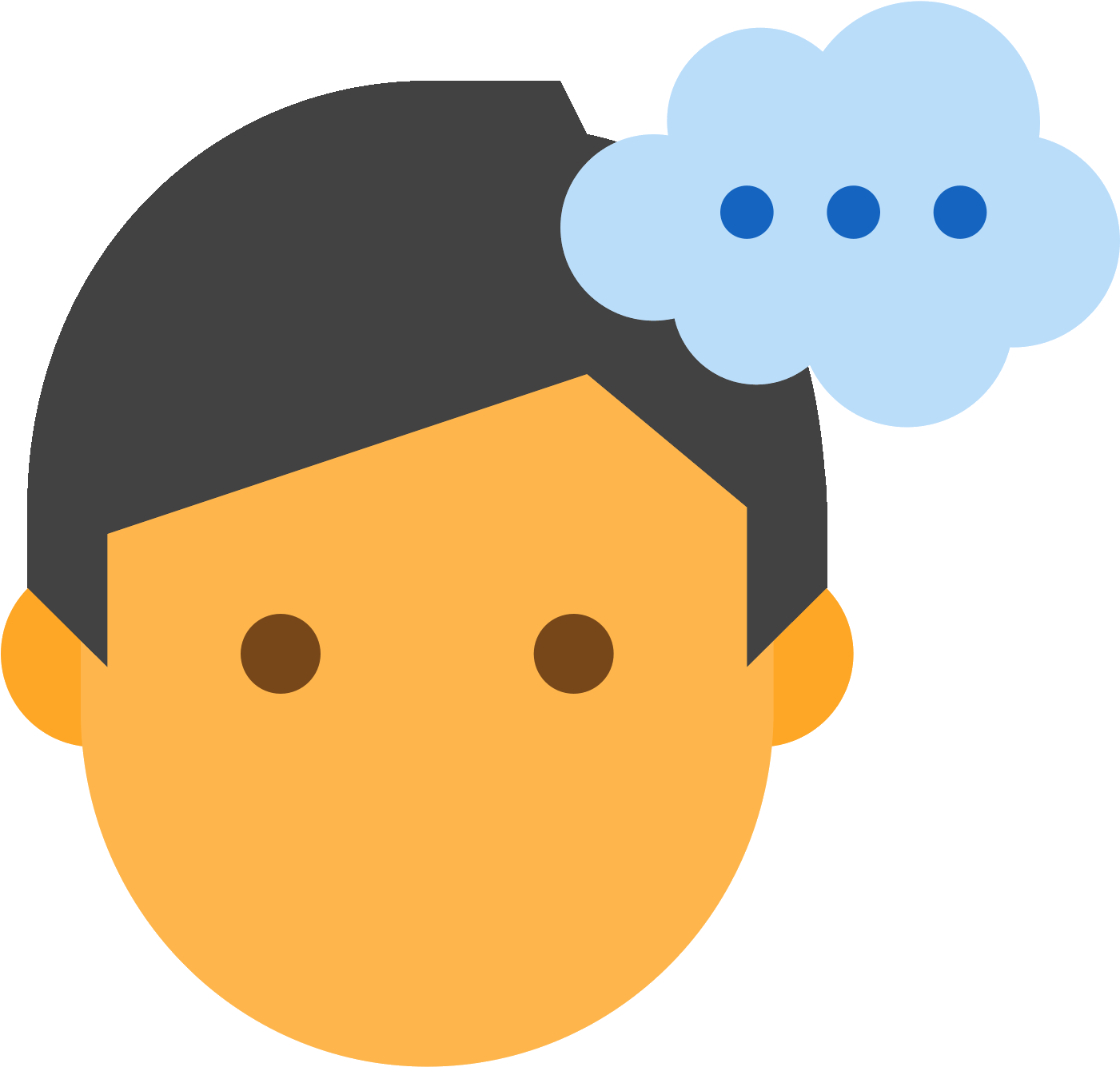}\,\,\,Tell me a bio of Bill Gates.},
		colback=gray!2!white,
		colframe=cyan!15!white,
            coltitle=black,
		width=32em,
		sharp corners=northwest,
		]
			\includegraphics[width=1.5em]{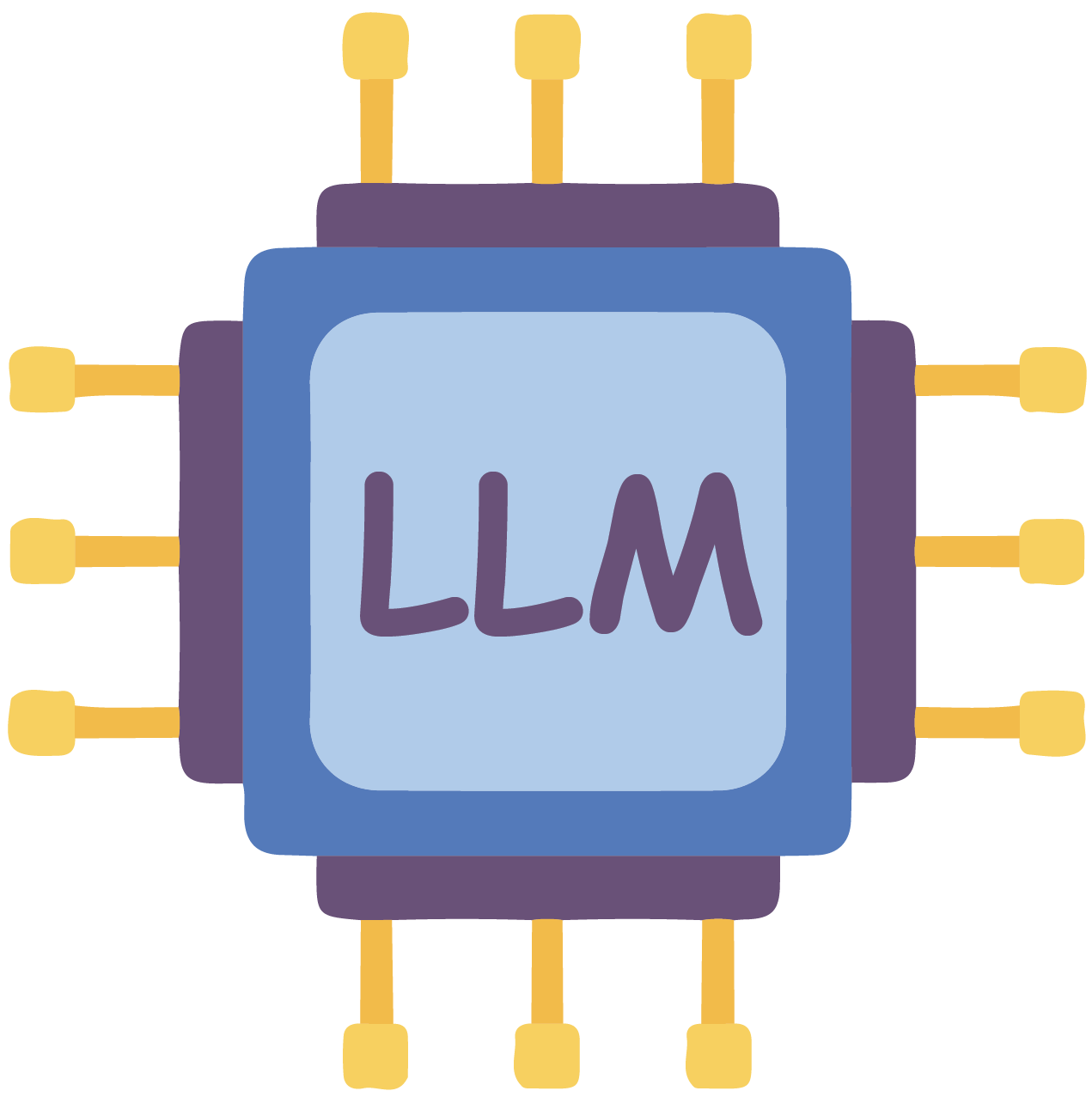}
			 \,\,\,Bill Gates, born William Henry Gates III on October 28, 1955, is an American entrepreneur, software developer, and philanthropist. He is best known as the co-founder of Microsoft Corporation, one of the world's largest and most successful software companies. His net worth is estimated to be around \$131.7 billion and is consistently ranked among the world's wealthiest people.
Gates \marktext{myorange}{grew up in Seattle, Washington}, with his \marktext{myorange}{parents, an attorney and a schoolteacher}, and \marktext{myorange}{two sisters}. He showed an early aptitude for mathematics and computer programming and attended Lakeside School, a private preparatory school in Seattle where he met and befriended his future business partner, Paul Allen. In 1973, Gates enrolled at Harvard University, but he soon found himself uninterested in his classes and spent much of his time on programming projects. He eventually dropped out of Harvard to focus on developing Microsoft.
\marktext{myorange}{Gates and Allen started Microsoft in 1975}, initially providing software for the Altair 8800, an early personal computer.

	\end{tcolorbox}

\columnbreak

Claim-Conditioned Probability (ours)

\begin{tcolorbox}[size=fbox,
		title=\,{\includegraphics[width=1.5em]{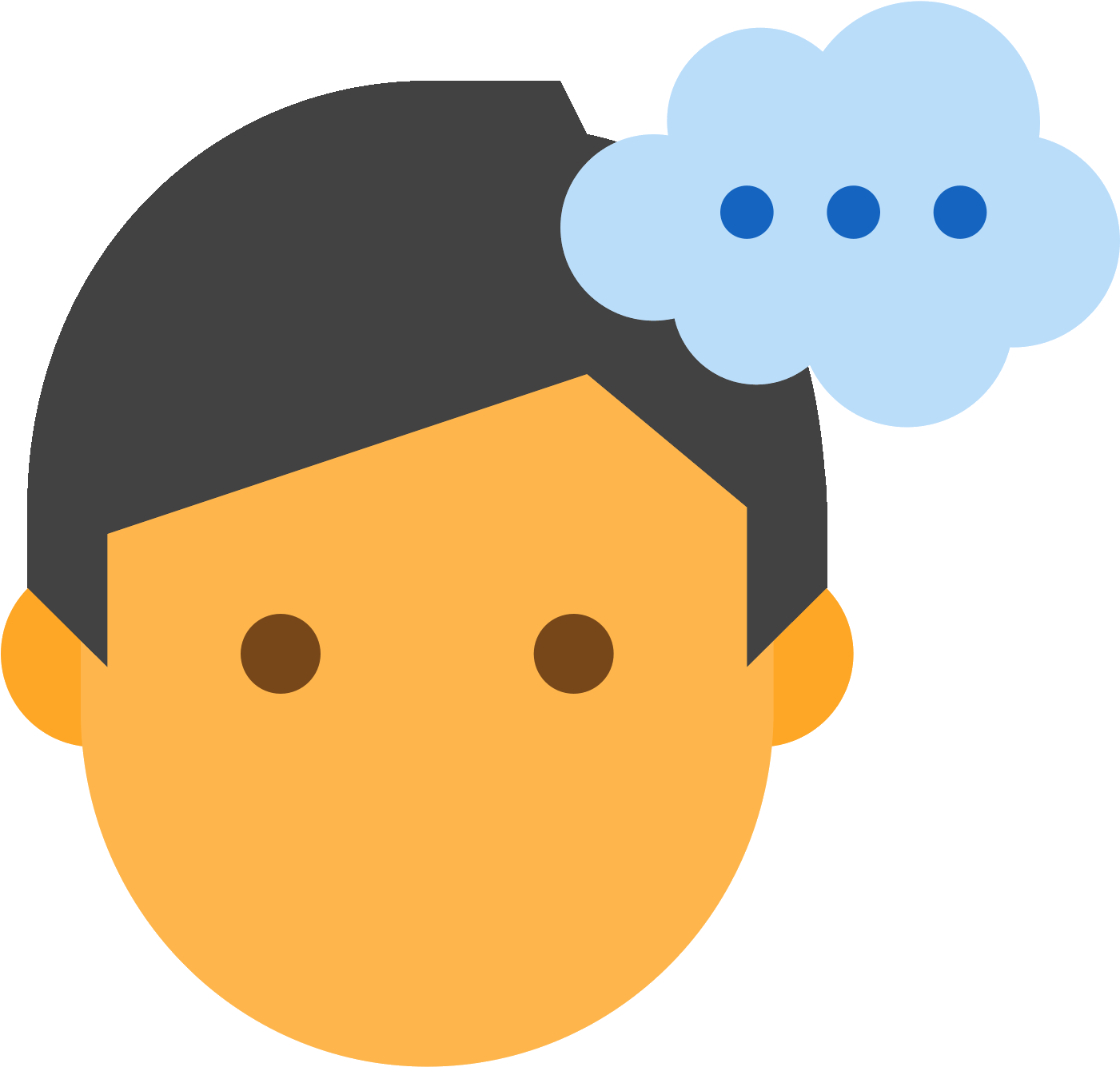}\,\,\,Tell me a bio of Bill Gates.},
		colback=gray!2!white,
		colframe=cyan!15!white,
            coltitle=black,
		width=32em,
		sharp corners=northwest,
		]
			\includegraphics[width=1.5em]{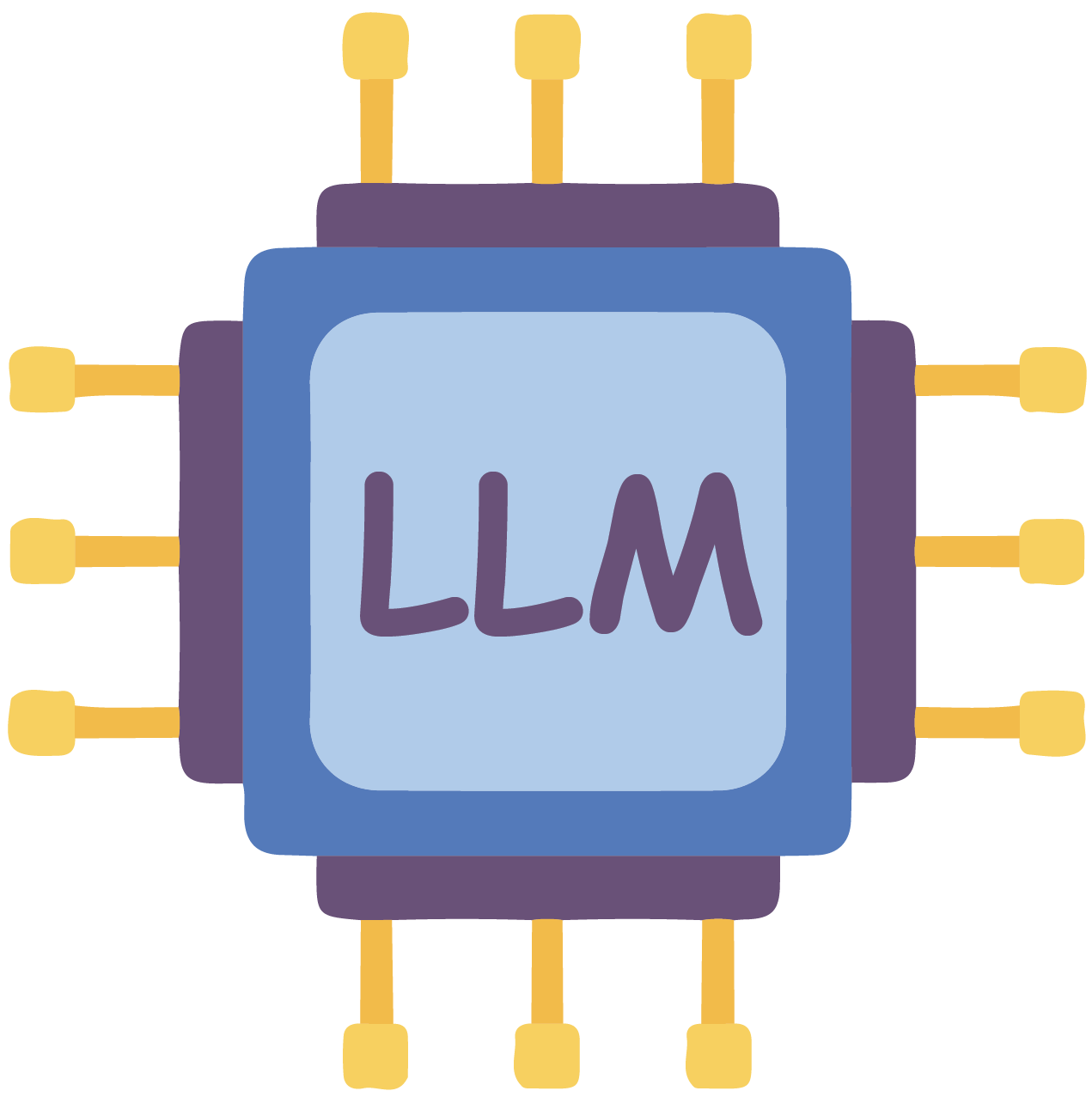}
			\,\,\,Bill Gates, born William Henry Gates III on October 28, 1955, is an American entrepreneur, software developer, and philanthropist. He is best known as the co-founder of Microsoft Corporation, one of the world's largest and most successful software companies. His net worth is estimated to be around \$131.7 billion and is consistently ranked among the world's wealthiest people.
Gates grew up in Seattle, Washington, with his parents, an attorney and a schoolteacher, and two sisters. He showed an early aptitude for mathematics and computer programming and attended Lakeside School, a private preparatory school in Seattle where he met and befriended his future business partner, Paul Allen. In 1973, Gates enrolled at Harvard University, but he soon found himself uninterested in his classes and spent much of his time on programming projects. He eventually dropped out of Harvard to focus on developing Microsoft.
Gates and Allen started Microsoft in 1975, initially providing software for the Altair 8800, an early personal computer.
	\end{tcolorbox}
 
 \end{multicols}
	    \caption{The Maximum Probability baseline vs Claim-Conditioned Probability (ours).}
	\label{fig:token-level_comparison}
\end{figure*}

  Figures~\ref{fig:token-level_comparison_app2}, \ref{fig:token-level_comparison_app} and~\ref{fig:token-level_comparison} show examples of the Claim-Conditioned Probability method compared with Maximum Probability. On all examples, our CCP method correctly detects the wrong facts, whereas Maximum Probability mistakenly highlights correct facts.

  \setlength{\columnsep}{0.3pc}

  \setlength{\columnsep}{0.3pc}

  \setlength{\columnsep}{0.3pc}

\end{document}